\documentclass[journal]{IEEEtran}
\usepackage{amsmath,amsfonts,amssymb,amsthm}
\usepackage{algorithmic}
\usepackage{algorithm}
\usepackage{array}
\usepackage[caption=false,font=normalsize,labelfont=sf,textfont=sf]{subfig}
\usepackage{textcomp}
\usepackage{stfloats}
\usepackage{url}
\usepackage{verbatim}
\usepackage{graphicx}
\usepackage{cite}
\usepackage{enumerate}
\usepackage{booktabs}
\usepackage{pifont}
\usepackage{multirow}
\usepackage{mathrsfs}
\usepackage{rotating}
\usepackage{bm}
\usepackage{wrapfig}
\usepackage[dvipsnames,svgnames,x11names]{xcolor}
\newcommand{\bftab}[1]{{\fontseries{b}\selectfont#1}}
\newcommand{\tabincell}[2]{\begin{tabular}{@{}#1@{}}#2\end{tabular}}

\newcommand{\ie}{\textit{i.e.}}

\begin{document}
\title{Attribute-Centric Vision-Language Alignment for Fine-Grained Gait Understanding}
\author{Saihui~Hou,~\IEEEmembership{Member,~IEEE,}
Chenye~Wang, Wenpeng~Lang, Qiannan~Guo, Zhengxiang~Lan,
Chao~Fan,~\IEEEmembership{Member,~IEEE,}
Shiqi~Yu,~\IEEEmembership{Member,~IEEE,}
Yongzhen~Huang,~\IEEEmembership{Senior Member,~IEEE}
\thanks{Saihui Hou, Qiannan Guo, and Zhengxiang Lan are with School of Artificial Intelligence, Beijing Normal University. (Email: housaihui@bnu.edu.cn, guoqiannan@mail.bnu.edu.cn, zhengxianglan@mail.bnu.edu.cn)}
\thanks{Chenye Wang is with The Hong Kong University of Science and Technology (Email: chenye.wang@connect.ust.hk). Wenpeng Lang is with Zhejiang University (Email: wenpenglang@zju.edu.cn). They contributed to this work while they were with Beijing Normal University.}
\thanks{Chao Fan is with School of Artificial Intelligence, Shenzhen University (Email: chaofan996@szu.edu.cn).}
\thanks{Shiqi Yu is with Department of Computer Science and Engineering at the Southern University of Science and Technology (Email: yusq@sustech.edu.cn).}
\thanks{Yongzhen Huang is with School of Artificial Intelligence, Beijing Normal University and also with Watrix Technology Limited Co. Ltd. (Email: huangyongzhen@bnu.edu.cn)}}

\maketitle
\begin{abstract}
Walking sequences reveal human-interpretable attributes related to body structure, posture, and motion.
Gait attribute recognition provides a concrete route to fine-grained gait understanding by predicting these attributes.
However, existing benchmarks provide limited fine-grained annotations, while classifier-based gait attribute recognition methods typically represent candidate classes through learned output weights rather than directly matching visual evidence with textual class representations.
To address the annotation gap, we present RA-GAR, a large-scale, richly annotated multimodal benchmark for gait attribute recognition, containing 1,001 subjects, 242,729 walking sequences, and annotations for 15 static and motion gait attributes.
We further propose Gait-VL, an attribute-centric vision-language alignment framework that recognizes gait attributes by matching attribute-specific multi-frame visual evidence with textual class prototypes.
On the visual side, attribute-specific learnable queries aggregate RGB features across spatial locations and frames into per-attribute visual embeddings.
On the language side, diverse descriptions expanded by a large language model are combined with context about relevant body regions to construct textual class prototypes.
Beyond attribute recognition, we evaluate the transferability of the attribute-pretrained visual encoder by using it to replace generic upstream large vision models in identity-oriented gait recognition pipelines; attribute pre-training uses no identity classification or metric-learning objectives.
Experiments demonstrate state-of-the-art attribute recognition performance on RA-GAR and consistent improvements in downstream identity recognition across multiple pipelines and benchmarks, supporting the effectiveness and transferability of attribute-centric gait representation learning.
\end{abstract}

\begin{IEEEkeywords}
Gait Attribute Recognition, Vision-Language Alignment, Semantic Pre-training
\end{IEEEkeywords}

\section{Introduction}
\label{sec:intro}

\begin{figure}[t]
	\centering
	\includegraphics[width=0.99\linewidth]{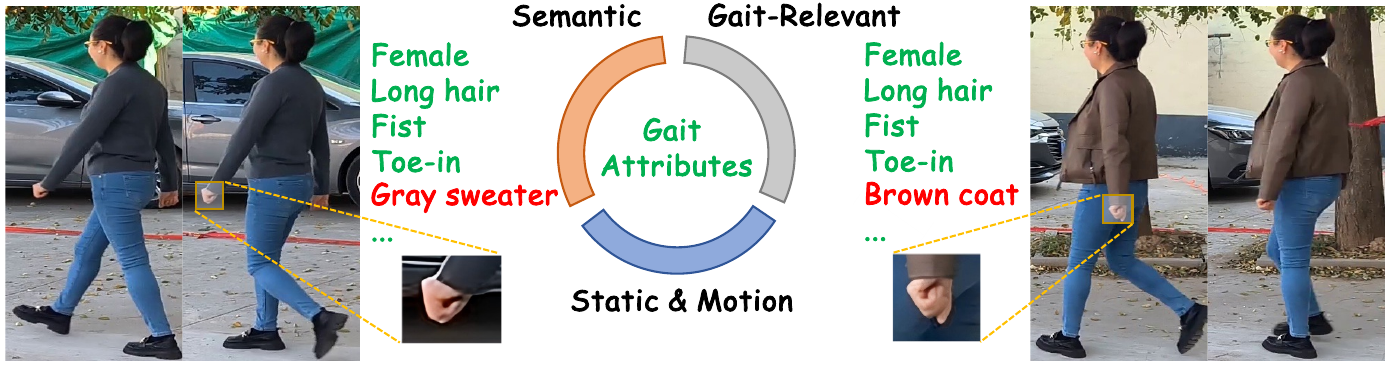}
	\caption{
		Illustration of the gait-attribute design, emphasizing semantic interpretability, gait relevance, and coverage of static and motion attributes. Green labels show representative attributes included in the taxonomy, whereas red labels show clothing color and garment-type cues excluded from it.
	}
	\label{fig:gar_definition}
\end{figure}

\IEEEPARstart{G}{ait} is a distinctive behavioral biometric because it can be observed at a distance without requiring active cooperation from the subject~\cite{bouchrika2011using}.
While gait identity recognition focuses on answering \textit{who} a person is, a walking sequence also reveals \textit{how} the person walks.
Such a sequence carries rich semantic cues about body structure, posture, and motion, such as toe orientation, arm swing, dragging feet, and limping, as illustrated in Fig.~\ref{fig:gar_definition}.
In this work, we address the attribute-level aspect of fine-grained gait understanding by recognizing these human-interpretable attributes from walking sequences.
The resulting attribute predictions could support retrieval when identity is unknown, provide semantic cues for studying abnormal walking patterns, and offer human-readable descriptions that complement identity-oriented gait analysis.

However, benchmarks dedicated to fine-grained gait attribute recognition remain limited.
Identity-oriented gait datasets have expanded substantially in scale~\cite{takemura2018multi,zhu2021gait}, yet even these resources provide only a few coarse annotations such as gender, age, or height~\cite{yu2006framework,zhu2021gait,zheng2022gait,song2022casia} rather than fine-grained gait attributes.
The closest dedicated benchmark, MA-Gait~\cite{song2023gait}, contains only 13,954 sequences and represents each of its 12 gait-oriented attributes using \emph{binary} labels, limiting both dataset scale and the coverage of meaningful multi-state gait semantics.
Other resources study adjacent problems, including age estimation~\cite{makihara2011gait,xu2017isir} and abnormal walking styles~\cite{ortells2018vision}, but do not target comprehensive gait attribute recognition.
Taken together, these limitations highlight the need for a large-scale benchmark with broad coverage of gait attributes and fine-grained annotations of their states.

Beyond these dataset limitations, existing gait attribute recognition methods typically predict attributes through task-specific classification heads~\cite{song2023gait}.
These heads represent each candidate class through learned output weights, without explicitly using textual class definitions to construct these weights.
Such definitions can describe the relevant body regions and attribute states, providing explicit semantic references for matching visual evidence to candidate classes.
We therefore seek to align attribute-specific visual representations learned from walking sequences with textual class representations.

To address these gaps and support fine-grained gait understanding at the attribute level, we introduce \textbf{RA-GAR}, a richly annotated benchmark for gait attribute recognition, and \textbf{Gait-VL}, an attribute-centric vision-language alignment framework.
We further examine whether attribute-level semantic supervision transfers to identity-oriented gait recognition.

On the data side, RA-GAR contains 1,001 subjects and 242,729 walking sequences with complete annotations for 15 gait attributes.
The taxonomy spans static characteristics and motion patterns, comprising 7 binary and 8 multi-class attributes with 45 candidate labels in total.
The benchmark is over an order of magnitude larger than MA-Gait~\cite{song2023gait} in both the numbers of subjects and sequences.
Under a shared annotation system, RA-GAR provides RGB, silhouette, and 2D/3D pose modalities across multiple viewpoints, with bag carrying and clothing changes included as walking covariates.

On the modeling side, Gait-VL formulates gait attribute recognition as a sequence-text matching problem.
Rather than using learned classifier weights as class representations, it constructs textual prototypes from banks of human-readable descriptions and measures their compatibility with walking evidence.
Unlike generic image-text matching, fine-grained gait attribute recognition requires associating language with evidence that can be subtle, body-localized, and distributed across multiple frames.
Gait-VL therefore adopts two complementary designs.
First, on the visual side, attribute-specific learnable queries aggregate RGB features across spatial locations and frames into per-attribute visual embeddings.
Second, on the textual side, Gait-VL constructs gait-specific textual class prototypes from description banks that combine descriptions expanded by a large language model (LLM) with context about relevant body regions.
Using multiple descriptions is intended to reduce sensitivity to any single wording while keeping the meaning of each candidate class explicit.
Together, these designs explicitly align attribute-aware sequence representations with human-readable gait semantics.

We further investigate whether learning these fine-grained gait semantics produces representations that transfer to identity-oriented gait recognition.
Recent methods have begun to use large-scale pretrained models as generic visual representation sources for gait recognition~\cite{ye2024biggait,ye2025biggergait}.
Generic visual pre-training does not explicitly organize representations around gait-specific factors.
Attribute-centric pre-training provides a complementary signal by organizing the visual encoder around body structure, posture, and multi-frame walking patterns without identity-level objectives.
By replacing generic upstream large vision models (LVMs) in recent gait recognition pipelines with the attribute-pretrained visual encoder from Gait-VL, we show that semantic gait attribute learning can improve downstream identity recognition, suggesting that gait attributes are not merely auxiliary labels but reusable supervision for gait representation learning.

In summary, the main contributions of this paper are:
\begin{itemize}
	\item We establish \textbf{RA-GAR}, a large-scale multimodal benchmark for fine-grained gait attribute recognition, comprising 1,001 subjects and 242,729 walking sequences. Under a shared annotation system spanning RGB, silhouette, and 2D/3D pose modalities, its taxonomy defines 15 static and motion gait attributes---7 binary and 8 multi-class---with 45 candidate labels in total.
	\item We propose \textbf{Gait-VL}, an attribute-centric vision-language framework that reformulates gait attribute recognition as sequence-text matching, where candidate labels are specified through language rather than learned classifier weights. Attribute-specific learnable queries aggregate RGB features across spatial locations and frames, while LLM-assisted expansion constructs description banks with context about relevant body regions for explicit visual-semantic alignment.
	\item We develop and evaluate an attribute-centric pre-training strategy for identity-oriented gait recognition. Using RA-GAR attributes as semantic supervision without identity classification or metric-learning objectives, the Gait-VL visual encoder replaces generic upstream large vision models in BigGait~\cite{ye2024biggait} and BiggerGait~\cite{ye2025biggergait}, while their downstream architectures, objectives, and evaluation protocols remain unchanged.
\end{itemize}

This paper substantially extends our conference version~\cite{wang2025ra} in benchmark construction, task formulation, model design, and evaluation.
The conference version introduced an early 533-subject subset of RA-GAR and addressed silhouette-based gait attribute recognition with CLIP-GAR, where predefined attribute descriptions were encoded by a CLIP text encoder and fused with silhouette-derived gait features for classifier-based prediction.
In contrast, this journal version adopts an attribute-centric vision-language formulation of gait attribute recognition and makes four major extensions.
First, we expand RA-GAR from the 533-subject conference dataset to a 1,001-subject benchmark with 242,729 sequences by incorporating 468 newly collected subjects, enabling a larger evaluation protocol and separate reporting for the two acquisition scenarios.
Second, we move from using textual descriptions mainly as auxiliary semantic features for classifier-based prediction to a sequence-text matching formulation, where textual label prototypes directly define the attribute prediction space.
Third, we replace the silhouette-based CLIP-GAR model with Gait-VL, a new RGB-based attribute-centric vision-language framework that explicitly integrates attribute-specific multi-frame aggregation, attribute-aware pooling, and LLM-assisted text expansion.
Finally, we extend the evaluation to semantic pre-training for identity-oriented gait recognition, showing that attribute-centric alignment can improve downstream recognition without relying on identity classification or metric-learning objectives during pre-training.
Together, these extensions broaden the conference study from text-assisted attribute classification to attribute-centric sequence-text matching and pre-training for gait identity recognition.

\section{Related Work}
\label{sec:related_work}

\subsection{Gait Datasets}

Public gait datasets have played a fundamental role in advancing gait analysis, with many representative benchmarks designed for identity recognition rather than semantic gait description.
Early controlled datasets, such as CASIA-B~\cite{yu2006framework}, established standard protocols for cross-view gait recognition under normal walking, bag-carrying, and coat-wearing conditions.
OU-MVLP~\cite{takemura2018multi} further enlarged the population scale and provided a large multi-view benchmark for silhouette-based gait recognition.
More recent datasets, including GREW~\cite{zhu2021gait}, Gait3D~\cite{zheng2022gait}, and CASIA-E~\cite{song2022casia}, have improved the scale, diversity, and realism of gait benchmarks by introducing unconstrained scenes, more subjects, or richer acquisition conditions.
Several benchmarks further focus on challenging covariates in practical gait recognition.
CCPG~\cite{li2023in} studies fine-grained clothes-changing conditions with RGB and silhouette sequences, SUSTech1K~\cite{shen2023lidargait} introduces LiDAR-based 3D gait data with diverse outdoor variations, and CCGR~\cite{zou2024cross} provides a large cross-covariate benchmark covering many views and walking conditions.
These datasets have greatly promoted identity-oriented gait recognition.
However, their annotations mainly describe identities, camera views, walking conditions, or a few coarse demographic attributes, leaving fine-grained body, posture, and motion concepts outside the primary evaluation target.

A complementary line of datasets considers gait-related semantic annotations.
The Whole Generation Gait Database~\cite{makihara2011gait} supports gait-based age estimation and shows that age-related semantic cues can be inferred from walking patterns.
OU-ISIR LP-Age~\cite{xu2017isir} further provides a large-population gait database with age annotations and protocols for gait-based age estimation.
The INIT Gait Database~\cite{ortells2018vision} supports vision-based gait impairment analysis by collecting typical and simulated pathological walking styles for aided diagnosis.
MA-Gait~\cite{song2023gait} is a representative benchmark for gait attribute recognition and introduces multiple static and motion attributes.
Together, these resources establish initial benchmarks for semantic gait analysis.
However, none of these datasets combines a large subject population, broad coverage of static and motion attributes, multiple data modalities, and fine-grained sequence-level annotations.
RA-GAR complements this landscape by making fine-grained static and motion gait attributes the primary annotation target and providing RGB, silhouette, and pose sequences under a common annotation system.

\subsection{Gait Attribute Recognition}

Gait attribute recognition provides a concrete route to fine-grained gait understanding by inferring human-interpretable properties from walking sequences.
Early studies mainly focused on soft biometric cues such as age, using gait templates, handcrafted representations, or gait energy images~\cite{lu2010gait,makihara2011gait}.
With the development of deep learning, convolutional models have been introduced to predict human attributes from gait energy images or silhouette-based representations~\cite{zhang2022gait}.
These methods demonstrate that gait contains semantic information beyond identity, but their attribute sets are usually small and dominated by coarse-grained cues.
More recent studies have moved from single or coarse attributes toward richer multi-attribute gait understanding.
MA-Gait~\cite{song2023gait} formulates gait attribute recognition as a multi-label classification task and proposes GAR-Net to jointly model static and motion attributes.
GAR-Net predicts attributes through learned classification heads, where candidate classes are represented by learned output weights without explicit reference to textual class definitions.

Beyond gait-specific methods, pedestrian attribute recognition provides relevant methodological context because it also associates human visual evidence with semantic attribute labels.
Representative advances include spatially localized attribute modeling~\cite{tang2019improving}, temporal aggregation across video frames~\cite{chen2019temporal,wang2024spatio}, and visual-language or prompt-based attribute prediction~\cite{cheng2022simple,zhu2023learning,wu2025enhanced}.
However, these methods primarily target appearance-oriented attributes in images or general surveillance videos, whereas fine-grained gait attributes rely more strongly on body-localized posture and motion evidence distributed across a walking sequence.
These requirements motivate matching textual label prototypes with attribute-specific visual evidence gathered across different body regions and multiple frames.

\subsection{Gait Identity Recognition}

Beyond their role as recognition targets, semantic gait attributes may also provide an intermediate supervisory signal for learning transferable gait representations.
Gait identity recognition provides a relevant downstream setting for examining this transferability, as it requires person-discriminative walking representations that are robust to view, clothing, carrying condition, and scene changes.
Existing methods can be broadly grouped according to their input representations into model-based and appearance-based approaches.

Model-based methods exploit skeleton, pose, or body-structure cues to reduce their dependence on appearance, including graph-based skeleton modeling~\cite{teepe2021gaitgraph}, generalized pose representations~\cite{fu2023gpgait}, physics-augmented 3D skeleton learning~\cite{guo2023physics}, dense-pose modeling~\cite{lang2025beyond}, and skeleton-map representations~\cite{fan2024skeletongait}.
These methods may inherit errors from upstream pose or skeleton estimation, particularly under low resolution or occlusion.
In contrast, appearance-based methods learn directly from silhouettes or RGB sequences, with representative directions including set and hybrid sequence modeling~\cite{chao2019gaitset,hou2025gaitasset,hou2025gaitsnippet}, part-based spatio-temporal modeling~\cite{fan2020gaitpart,lin2021gait,wang2023hierarchical,ma2024learning}, and robustness- or causality-oriented learning~\cite{huang2024occluded,dou2023gaitgci,xiong2024causality}.
Despite their different input representations, both families are primarily optimized for identity discrimination.

A closely related line of person re-identification (ReID) research studies semantic supervision for identity-oriented representation learning.
These methods use attributes to improve or interpret identity representations~\cite{lin2019improving,chen2021explainable,huang2024attribute}, typically through joint attribute--identity objectives or attribute-based model interpretation.
However, these studies do not isolate attribute-only pre-training.
To assess the transferable value of attribute supervision, we pre-train the visual encoder with attribute objectives alone and evaluate its transfer to downstream gait identity recognition.

\begin{table*}[t]
	\caption{Attribute taxonomy and class-level annotation labels of RA-GAR. The 15 attributes include 7 binary and 8 multi-class attributes, with 45 candidate labels in total.}
	\label{table:gait_attributes}
	\centering
	\setlength{\tabcolsep}{6pt}
	\renewcommand{\arraystretch}{1.1}
	\resizebox{0.99\textwidth}{!}{%
	\begin{tabular}{c|c|c|l|l}
		\toprule
		Type & Region & Attr. ID & Attribute & Class Labels (Index: Label) \\ \hline
		\multirow{9}{*}{Static} & \multirow{4}{*}{Full body} & A1 & Gender & 0: male; 1: female \\
		& & A2 & Age & 0: \(<18\); 1: 18--60; 2: \(>60\) \\
		& & A3 & Height & \tabincell{l}{0: \(<150\) cm; 1: 150--159 cm;\\2: 160--169 cm; 3: 170--179 cm; 4: \(\geq 180\) cm} \\
		& & A4 & Body mass & 0: underweight; 1: normal; 2: overweight; 3: obesity \\ \cline{2-5}
		& \multirow{2}{*}{Head} & A5 & Hair & 0: short hair; 1: long hair; 2: bald; 3: hat \\
		& & A6 & Forward head position & 0: no; 1: yes \\ \cline{2-5}
		& \multirow{6}{*}{Upper body} & A7 & Uneven shoulders & \tabincell{l}{0: left shoulder higher; 1: right shoulder higher;\\2: even shoulders} \\
		& & A8 & Humpback & 0: no; 1: yes \\
		& & A9 & Hands & 0: natural; 1: fist; 2: asymmetrical; 3: others \\ \cline{1-1}\cline{3-5}
		\multirow{6}{*}{Motion} & & A10 & Arm swing direction & \tabincell{l}{0: inward towards the body; 1: outward away from the body;\\2: parallel to the body; 3: without arm swing; 4: single arm swing} \\
		& & A11 & Arm swing amplitude & 0: large; 1: small \\
		& & A12 & Upper arm movement & 0: no; 1: yes \\ \cline{2-5}
		& \multirow{3}{*}{Lower body} & A13 & Toe & 0: toe-in; 1: toe-out; 2: normal \\
		& & A14 & Drag feet & 0: no; 1: yes \\
		& & A15 & Limping & 0: no; 1: yes \\ \bottomrule
	\end{tabular}}
\end{table*}

\subsection{Vision-Language Learning and Gait Pre-training}

Vision-language learning and representation pre-training provide relevant foundations both for associating fine-grained gait attributes with textual semantics and for learning transferable gait representations.
Image-text foundation models such as CLIP~\cite{radford2021learning} and SigLIP2~\cite{tschannen2025siglip} provide aligned image-text encoders that support text-prompt-based semantic recognition.
Prompt learning adapts such models by optimizing textual context for downstream categories~\cite{zhou2022learning}, while language-model-generated descriptions enrich category prompts with discriminative semantics~\cite{pratt2023does}.
However, these paradigms operate primarily at the image-text level and do not directly associate gait sequences with fine-grained gait attribute descriptions.

Video, action, and motion pre-training methods move beyond static images and better capture temporal structure.
VideoMAE~\cite{tong2022videomae} demonstrates the effectiveness of masked video modeling for self-supervised video representation learning.
ActionCLIP~\cite{wang2021actionclip} reformulates action recognition as video-text matching.
MotionBERT~\cite{zhu2023motionbert} learns transferable human motion representations from large-scale motion data.
These methods provide useful temporal, action, or motion representations, but they do not use fine-grained gait attribute descriptions as sequence-level training targets.

Several studies have also explored pre-training for person ReID.
LUPerson~\cite{fu2021unsupervised} learns person representations from large-scale unlabeled pedestrian images.
PASS~\cite{zhu2022pass} introduces part-aware self-supervised pre-training for person ReID.
SOLIDER~\cite{chen2023beyond} learns semantic-controllable representations for human-centric visual tasks.
CLIP-ReID~\cite{li2023clip} further adapts vision-language models to image ReID without concrete text labels.
These methods focus on image-level person representations, whereas our work investigates explicit gait attribute supervision for learning transferable sequence representations.

Gait-specific representation learning further narrows the gap between generic pre-training and downstream gait analysis.
GaitSSB~\cite{fan2023learning} learns gait representations from massive unlabeled walking videos, while BigGait~\cite{ye2024biggait} and BiggerGait~\cite{ye2025biggergait} adapt features from large vision models to gait recognition.
More recently, language-oriented models have been introduced into gait identity recognition.
GaitLLM~\cite{yang2025bridging} maps gait tokens into and out of an LLM feature space to exploit its sequence-modeling capability.
Gait-World~\cite{huang2025vocabulary} uses gait-cycle vocabularies from vision-language models to guide identity-discriminative representation learning.
LMGait~\cite{wu2026language} further introduces natural-language descriptions as semantic priors together with motion-aware temporal modeling.
Although these approaches demonstrate the benefits of gait-domain pre-training and language guidance, they focus on identity-oriented representations rather than treating fine-grained gait attributes as sequence-level prediction targets.

\section{Datasets}
\label{sec:datasets}

This section presents RA-GAR, a large-scale benchmark for fine-grained gait attribute recognition.
Unlike identity-oriented gait datasets, RA-GAR treats semantic walking attributes as the primary target.
The dataset contains 1,001 subjects and 242,729 sequences, with annotations for 15 gait attributes covering static visual characteristics and dynamic walking patterns.
It spans diverse acquisition conditions, viewpoints, and walking covariates, including bag carrying and clothing changes.
These annotations support the study of fine-grained gait understanding through attribute recognition.

\subsection{Gait Attribute Definition}

We define gait attributes with three requirements in mind.
First, \textbf{each attribute should carry clear semantic meaning.}
Rather than using low-level statistical descriptors, we focus on human-understandable concepts that can be described in language and interpreted from visual evidence.
Second, \textbf{each attribute should describe a gait-relevant characteristic observable from walking sequences.}
In this work, we focus on body structure, posture, and walking patterns, leaving emotional states outside the annotation scope.
We exclude clothing color, texture, and garment-type cues commonly considered in pedestrian or re-identification attribute recognition~\cite{wang2022pedestrian}, while retaining head appearance through the hair and headwear categories in our taxonomy.
Third, \textbf{the attribute set should cover both static visual characteristics and dynamic walking patterns}, because gait semantics arise jointly from body structure, posture, and motion.

Based on these requirements, RA-GAR organizes gait attributes into a two-level taxonomy.
At the semantic level, the attributes are divided into static attributes and motion attributes.
Static attributes describe body, appearance, and posture characteristics, including full-body properties, head-related appearance, and upper-body posture states.
Motion attributes describe dynamic walking patterns, such as arm swing, toe orientation, dragging feet, and limping.
At the body-region level, attributes are grouped into full-body, head, upper-body, and lower-body categories according to their primary visual cues.
This grouping organizes the attributes; it does not define region-level annotations or constrain visual feature aggregation to predefined body regions.
In total, RA-GAR provides 15 gait attributes, including 7 binary attributes and 8 multi-class attributes.
Table~\ref{table:gait_attributes} summarizes the attribute taxonomy and the class-level annotation labels.

To reduce ambiguity in label assignment, we define explicit rules for selected attributes.
For body mass (A4), body mass index (BMI) is computed as weight in kilograms divided by squared height in meters, and thresholds of 18.5, 25, and 30 are used to define four operational categories~\cite{who2000obesity}.
For posture- and hand-related attributes, the labels are assigned according to visible body states during walking: forward head position (A6) indicates a visible forward-head posture, while hands (A9) distinguishes natural hands, fists, asymmetric hand posture, and other cases such as hands behind the back or in pockets.
For motion attributes, the labels are determined from sequence-level walking evidence.
Motivated by gait biomechanics studies that quantify arm-swing amplitude during walking~\cite{hill2019effects}, we adopt 30\({}^{\circ}\) as an operational annotation threshold for A11, labeling the amplitude as large when the swing exceeds this threshold and small otherwise.
Drag feet (A14) records whether the feet tend to scrape or drag without clear lifting, and limping (A15) records whether the walking pattern relies more on one leg than the other.
Together, these rules specify how labels are assigned from body measurements and from visible cues of posture and motion.

\subsection{Data Collection}

\begin{figure}[t]
	\centering
	\includegraphics[width=0.99\linewidth]{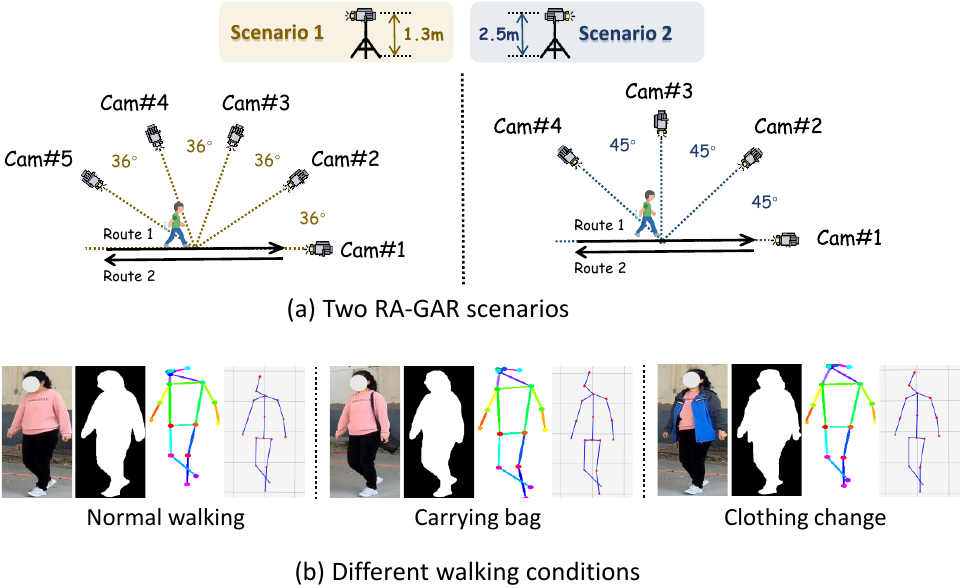}
	\caption{(a) Camera configurations and walking routes of the two RA-GAR scenarios.
	(b) Examples of RGB images, silhouettes, 2D poses, and 3D poses under normal walking, bag carrying, and clothing changes.
	}
	\label{fig:ragar_platform}
	\label{fig:ragar_visualization}
\end{figure}

RA-GAR was collected in two scenarios with different acquisition conditions, as illustrated in Fig.~\ref{fig:ragar_platform}.
The two scenarios share the same attribute annotation system and walking-condition design but use different capture configurations.

Scenario 1 contains 533 subjects and uses an outdoor acquisition setup with natural illumination variation.
Five physical cameras are fixed at a height of 1.3 meters.
Subjects walk along the route in both directions, yielding 10 effective views uniformly distributed over 360\({}^{\circ}\) at 36\({}^{\circ}\) intervals.
Videos were recorded at a resolution of 1920\(\times\)1080 and 30 FPS.
Each subject completes three attribute states: one natural state and two instructed attribute combinations.
For each attribute state, subjects complete four walks without bag or clothing-change covariates, two while carrying a bag, and two after changing clothes.
Thus, each subject in the first scenario is expected to provide 240 sequences before quality filtering, \ie, 10 views \(\times\) \((1+2)\) attribute states \(\times\) \((4+2+2)\) repetitions.

Scenario 2 contains 468 subjects and follows the same attribute annotation system and walking-condition design under a different acquisition configuration.
It differs from Scenario 1 in background, illumination, camera placement, and view distribution.
Specifically, four cameras are fixed at a height of 2.5 meters with 45\({}^{\circ}\) angular intervals.
Subjects walk along the route in both directions, so the four physical cameras provide eight effective views.
Each subject completes four attribute states: one natural state and three instructed attribute combinations.
Therefore, each subject in the second scenario is expected to provide 256 sequences before quality filtering, \ie, 8 effective views \(\times\) \((1+3)\) attribute states \(\times\) \((4+2+2)\) walking-condition repetitions.

The instructed attribute combinations used in both scenarios were introduced to broaden attribute-state coverage and mitigate class imbalance.
Some gait attributes, such as limping, dragging feet, or pronounced postural patterns, occur infrequently in natural walking; passive collection alone would therefore provide relatively few positive samples.
To keep the resulting instructed states plausible, the attribute combinations are customized according to each subject's body shape, age, posture, and natural walking characteristics.
For example, elderly subjects are more likely to be assigned a humpback state, while foot-related motion states are selected to suit each subject's natural gait.
Before recording, subjects receive instructions and practice the assigned attribute combinations.
During both practice and recording, three annotators jointly verify that the attribute labels match the observed posture and motion.
In the RA-GAR annotations, gender, age, height, and body mass labels are shared across the collected sequences of each subject, whereas the posture and motion labels of the instructed states are assigned to each sequence according to its observable gait patterns.

\subsection{Data Processing and Statistics}

After data collection, the raw videos are processed into sequence-level samples through three steps.
First, the raw videos are manually segmented according to identity, view, walking condition, and attribute state.
Second, ByteTrack is applied to the segmented videos to track the target subject and generate subject-centered crops and bounding boxes~\cite{zhang2022bytetrack}.
Third, sequences with obvious splitting errors, incomplete walking trajectories, or severe tracking failures are removed before modality extraction.
The remaining sequences form the cleaned RA-GAR samples, each associated with a subject, scenario, view, walking condition, and a complete set of attribute annotations.

Based on the cleaned clips, RA-GAR provides multiple data modalities to support different research settings, as illustrated in Fig.~\ref{fig:ragar_visualization}.
In addition to the RGB clips, silhouette sequences are extracted with PaddleSeg~\cite{liu2021paddleseg}.
For pose-based research, ViTPose~\cite{xu2022vitpose} is used to estimate 2D keypoints, and MotionBERT~\cite{zhu2023motionbert} is used to recover 3D pose sequences.
These derived modalities support appearance-based, silhouette-based, and pose-based gait attribute analysis with shared annotations.

\begin{figure}[t]
	\centering
	\includegraphics[width=0.99\linewidth]{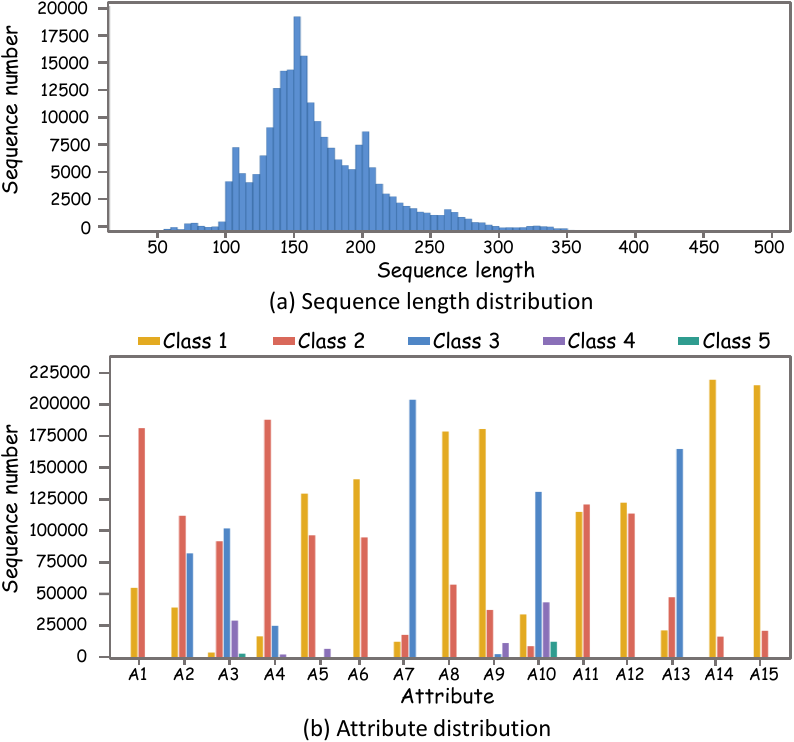}
	\caption{RA-GAR statistics: (a) distribution of sequence lengths measured in frames; (b) sequence counts for the classes of each attribute.
	A1--A15 correspond to the attributes in Table~\ref{table:gait_attributes}; Classes 1--5 follow the label order in that table, corresponding to indices 0--4.
	}
	\label{fig:ragar_statistics}
\end{figure}

After processing, RA-GAR contains 1,001 subjects and 242,729 valid video sequences collected from two scenarios.
The first scenario contains 533 subjects and 122,921 valid video sequences after quality filtering, while the second scenario contains 468 subjects and 119,808 valid video sequences.
Each valid sequence is labeled for all 15 gait attributes, including 7 binary attributes and 8 multi-class attributes.
Fig.~\ref{fig:ragar_statistics} visualizes the sequence-length distribution and attribute distribution.
Despite the targeted collection of instructed attribute states, several fine-grained attributes remain imbalanced because imposing rare states on every subject would reduce walking plausibility and annotation reliability.
We retain the resulting empirical distribution and evaluate performance under class imbalance using attribute-based mean accuracy together with standard instance-based metrics~\cite{huang2019deep}.

Table~\ref{table:datasets} compares RA-GAR with representative identity-oriented and gait-semantic datasets.
Among the compared gait-semantic benchmarks, RA-GAR uniquely combines fine-grained attribute annotations with a dedicated clothing-change protocol, supporting the evaluation of attribute recognition under appearance variations.

\begin{table*}[t]
	\caption{Comparison between RA-GAR and representative gait datasets. \#Subj., \#Seq., and \#Cam. denote the numbers of subjects, sequences, and physical cameras, respectively. The rows above and below the horizontal separator list identity-oriented and gait-semantic benchmarks, respectively. A checkmark and a cross indicate the presence and absence of a dedicated within-subject clothing-change protocol, respectively. For MA-Gait, only the 12 non-emotional gait-oriented attributes are counted. The effective view counts of OU-MVLP, CASIA-E, MA-Gait, and the two RA-GAR scenarios are 14, 26, 12, and 10/8, respectively. Slash-separated RA-GAR values refer to Scenario 1 and Scenario 2, respectively.}
	\label{table:datasets}
	\centering
	\normalsize
	\setlength{\tabcolsep}{6pt}
	\renewcommand{\arraystretch}{1.1}
	\resizebox{0.99\textwidth}{!}{%
	\begin{tabular}{c|ccccccl}
		\toprule
		Dataset & Year & \#Subj. & \#Seq. & \#Cam. & \tabincell{c}{Cloth-\\Changing} & \tabincell{c}{Gait-Oriented Attributes} & Modalities \\ \hline
		CASIA-B~\cite{yu2006framework} & 2006 & 124 & 13,640 & 11 & \ding{51} & Gender, Height & RGB, Silhouette  \\
		OU-MVLP~\cite{takemura2018multi} & 2018 & 10,307 & 288,596 & 7 & \ding{55} & -- & Silhouette \\
		GREW~\cite{zhu2021gait} & 2021 & 26,345 & 128,671 & 882 & \ding{55} & Age, Gender & \tabincell{l}{Silhouette, Optical Flow, 2D/3D Pose} \\
		Gait3D~\cite{zheng2022gait} & 2022 & 4,000 & 25,309 & 39 & \ding{55} & -- & \tabincell{l}{Silhouette, 2D/3D Pose, 3D Mesh} \\
		CASIA-E~\cite{song2022casia} & 2023 & 1,014 & 778,752 & 8 & \ding{51} & \tabincell{l}{Age, Gender, Height, Weight} & \tabincell{l}{Silhouette, Thermal IR} \\
		\hline
		OU-ISIR LP-Age~\cite{xu2017isir} & 2017 & 63,846 & 63,846 & 1 & \ding{55} & Age, Gender & Silhouette \\
		INIT Gait Database~\cite{ortells2018vision} & 2018 & 10 & 160 & 1 & \ding{55} & 8 Gait Styles & Silhouette \\
		MA-Gait~\cite{song2023gait} & 2024 & 95 & 13,954 & 6 & \ding{55} & 12 Binary & Silhouette, 2D Pose \\
		\textbf{RA-GAR} & \textbf{2026} & \textbf{1,001} & \textbf{242,729} & \textbf{5/4} & \ding{51} & \textbf{\tabincell{l}{7 Binary, 8 Multi-Class}} & \textbf{\tabincell{l}{RGB, Silhouette, 2D/3D Pose}} \\
		\bottomrule
	\end{tabular}}
\end{table*}

\subsection{Dataset Split and Evaluation Settings}
\label{sec:ragar_split}

RA-GAR uses a subject-disjoint partition, with all sequences from the same subject assigned to either training or testing.
The training split contains 500 subjects, including 250 subjects from each acquisition scenario.
The remaining 501 subjects form the full test set, comprising 283 subjects from Scenario 1 and 218 subjects from Scenario 2.
The two scenario-defined test subsets are disjoint and together constitute the full test set.
All available sequences from the test subjects, including sequences in their natural and instructed attribute states, are included in evaluation.
We use the full test set as the primary benchmark and report results separately for the two scenarios to examine performance differences.

Under this split, RA-GAR supports both classifier-based attribute prediction and text-guided sequence-text matching.
Both settings are evaluated under the same protocol, comprising 15 attributes and 45 candidate labels shared by training and testing.
The attribute annotations are also used as a semantic pre-training signal without identity classification or metric-learning objectives to evaluate transfer to downstream gait identity recognition.

\subsection{Consent, Privacy, and Data Access}

Before data collection, participants or their legal guardians, as applicable, provided written informed consent for data collection, processing, attribute annotation, and sharing for research under controlled access.
Requests for withdrawal and deletion of participant data can be submitted to the research team.
Each participant was assigned a pseudonymous dataset identifier, and names, contact information, and other direct administrative identifiers are not included in the released data.

After publication, RA-GAR will be made available through request-based access rather than unrestricted public download.
Applicants will be required to provide their institutional affiliation and intended research purpose.
The research team will review each application, and access will be granted only after approval and acceptance of the research-only data-use terms.
These terms will prohibit attempts to recover participants' real-world identities or link dataset records to external personal information, as well as unauthorized redistribution and non-research use.
Access may be revoked for violations.

\section{Methods}
\label{sec:methods}

To recognize the gait attributes defined above, we present Gait-VL, an attribute-centric vision-language framework that aligns RGB gait sequences with textual class semantics.
We first formalize the task and outline the pipeline, then detail its gait-centric backbone, sequence encoder, text encoder, matching objectives, and inference procedure.
Finally, we explain how its attribute-pretrained visual encoder replaces the generic upstream visual encoder in downstream gait recognition without identity-level pre-training losses.

\subsection{Problem Formulation}

We study supervised recognition of a predefined set of semantic gait attributes.
Let \(\mathcal{A}=\{a_m\}_{m=1}^{M}\) denote the set of \(M\) gait attributes.
Each attribute \(a_m\) is associated with \(N_m\) candidate labels, forming the label set \(\mathcal{L}_m=\{\ell_m^n\}_{n=0}^{N_m-1}\).
Let \(\mathcal{C}=\{(m,n)\mid m=1,\ldots,M,\ n=0,\ldots,N_m-1\}\) denote the complete candidate-label index set, whose cardinality is \(C=|\mathcal{C}|=\sum_{m=1}^{M}N_m\).
For example, the attribute \textit{Toe} contains labels such as toe-in, toe-out, and normal, while the attribute \textit{Limping} contains binary labels indicating whether limping appears.
Given an annotated gait sequence \(x_i\), let \(c_{i,m}\in\{0,\ldots,N_m-1\}\) denote the ground-truth class index of attribute \(a_m\), corresponding to label \(\ell_m^{c_{i,m}}\).
For candidate-level supervision, this categorical annotation is equivalently expanded into
\begin{equation}
	b_{i,m,n}=\mathbb{I}\!\left[c_{i,m}=n\right],
\end{equation}
where \(\mathbb{I}[\cdot]\) denotes the indicator function.

For sequence \(x_i\), the model produces a score vector
\begin{equation}
		\mathbf{s}_{i,m}=[s_{i,m,0},\ldots,s_{i,m,N_m-1}]\in\mathbb{R}^{N_m},
\end{equation}
over the candidate labels of each attribute \(a_m\in\mathcal{A}\).
Conventional gait attribute recognition usually realizes this mapping with classification heads over the predefined label sets.
Our formulation instead specifies candidate labels through language and predicts attributes by comparing gait sequences with textual descriptions of these labels.

\begin{figure*}[t]
	\centering
	\includegraphics[width=0.99\linewidth]{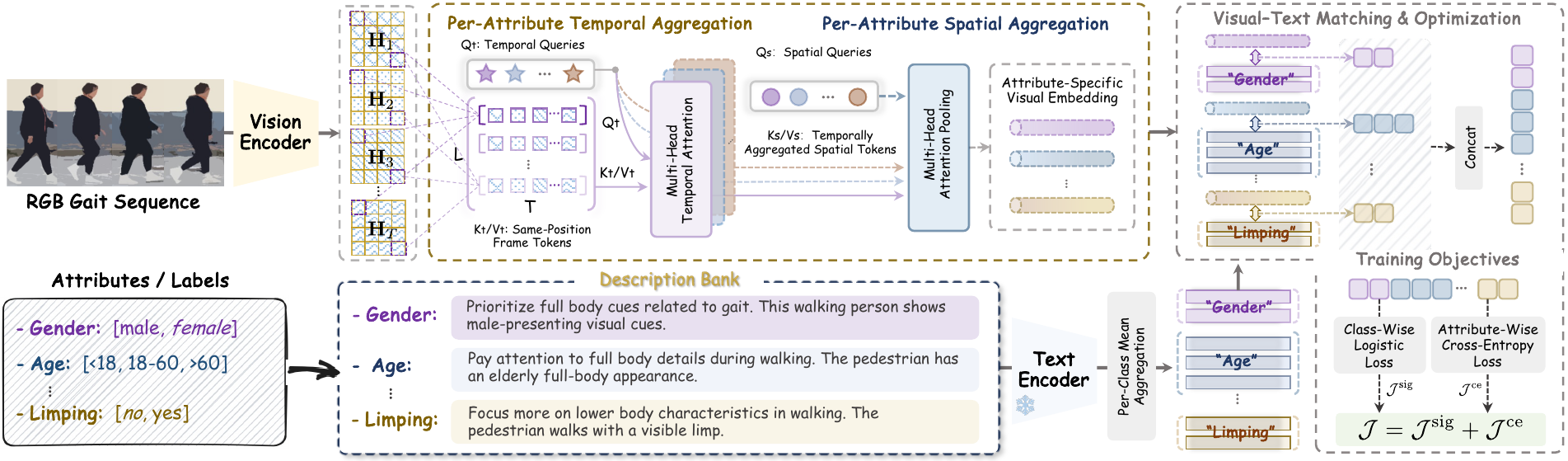}
	\caption{Overview of Gait-VL. SigLIP2 encodes RGB frames into spatial tokens. Attribute-specific temporal queries aggregate multi-frame evidence, after which attribute-specific spatial queries produce per-attribute visual embeddings through attention pooling. Here, \(Q_t\) and \(Q_s\) denote the query banks \(\{\mathbf{q}_m\}_{m=1}^{M}\) and \(\{\mathbf{p}_m\}_{m=1}^{M}\), respectively; \(K_t,V_t\) and \(K_s,V_s\) denote the keys and values in the corresponding temporal and spatial stages. On the text side, LLM-expanded descriptions are combined with context about relevant body regions and encoded into cached embedding banks. During training, sampled descriptions form stochastic textual prototypes that supervise sequence-text matching through class-wise sigmoid and attribute-wise softmax objectives; during inference, each complete embedding bank is averaged into a deterministic textual prototype.}
	\label{fig:gait_vl}
\end{figure*}

\subsection{Overview of Gait-VL}

Fig.~\ref{fig:gait_vl} illustrates how Gait-VL instantiates this text-based prediction interface through sequence-text matching.
For each attribute--label pair \((a_m,\ell_m^n)\), Gait-VL constructs a bank of complete descriptions from a template seed through LLM-assisted expansion.
This construction bridges the discrete attribute--label space and the language space while preserving the semantics of every predefined RA-GAR class.

Given an RGB gait sequence, the sequence encoder retains spatially resolved visual features, uses attribute-specific temporal queries to aggregate multi-frame evidence, and applies attribute-specific spatial queries to extract one visual embedding for each evaluated attribute.
On the text side, the attribute text encoder maps candidate-label descriptions into cached description embeddings, from which textual prototypes are constructed.
The matching module computes each element of \(\mathbf{s}_{i,m}\) by comparing the corresponding attribute-specific visual embedding with a candidate textual prototype.
Training uses stochastic prototypes constructed from sampled descriptions to improve semantic robustness, whereas inference aggregates the complete description bank into deterministic prototypes.

\subsection{Gait-Centric Vision-Language Backbone}

We consider three factors when selecting a backbone for the sequence-text matching formulation described above.
\textbf{(1) Cross-modal alignment.} Visual evidence and candidate-label descriptions need to be represented in a shared semantic space for matching.
\textbf{(2) Aspect-ratio preservation.} The backbone should accommodate portrait-shaped pedestrian crops to help preserve body proportions and limb configurations that provide evidence for attribute recognition.
Directly stretching such crops to a square can alter these cues, while square cropping can remove relevant body regions.
\textbf{(3) Encoding cost.} The cost of frame-level encoding is a practical consideration because each gait sequence contains multiple frames.

These considerations motivate our use of SigLIP2-Base-Patch16-NaFlex~\cite{tschannen2025siglip}.
Its image-text pre-training provides a cross-modal initialization for semantic matching, whereas purely visual foundation models such as DINOv2~\cite{oquab2023dinov2} and SAM~\cite{kirillov2023segment} do not natively provide a shared image-text embedding space.
The selected backbone is also adapted during pre-training to varying aspect ratios and spatial token grids, enabling portrait inputs without square resizing.
We use a compact model configuration to limit multi-frame encoding costs.
Within each sequence, frames are resized to a fixed portrait resolution, providing a consistent spatial token grid for the temporal aggregation described next.

\subsection{Gait Sequence Encoder}

The gait sequence encoder converts an RGB walking sequence into one visual embedding per attribute for sequence-text matching.
It retains frame-level spatial tokens, aggregates evidence across frames at each token position, and then pools the resulting tokens with attribute-specific spatial queries.

\subsubsection{Frame-Level Token Representation}

Given a gait sequence \(x=\{I_f\}_{f=1}^{T}\), where \(T\) is the number of frames and \(I_f\) denotes the \(f\)-th RGB pedestrian crop, each frame is resized to a portrait resolution and encoded by the visual encoder of SigLIP2.
For clarity, we omit the sequence index in this subsection.
We retain the post-layer-normalized spatial tokens before image-level pooling:
\begin{equation}
	\mathbf{H}_f = E_{\mathrm{vis}}(I_f), \quad \mathbf{H}_f \in \mathbb{R}^{L \times d},
\end{equation}
where \(\mathbf{H}_f=[\mathbf{h}_{f,1},\ldots,\mathbf{h}_{f,L}]^{\top}\), \(\mathbf{h}_{f,j}\in\mathbb{R}^{d}\) is the \(j\)-th visual token, \(L\) is the number of spatial tokens, and \(d\) is the feature dimension.
Here, \((\cdot)^{\top}\) denotes transposition.
Keeping token-level features is important because gait attributes are spatially structured: some attributes are reflected by the head and shoulders, while others depend on arms, feet, or the full body.

\begin{table*}[t]
	\caption{Template seeds and one representative LLM-expanded description for each selected class. The expansion process jointly introduces linguistic diversity and attribute-relevant gait context.}
	\label{tab:text_examples}
	\centering
	\setlength{\tabcolsep}{8pt}
	\renewcommand{\arraystretch}{1.1}
	\resizebox{0.99\textwidth}{!}{%
	\begin{tabular}{ccll}
		\toprule
		ID & Class & Template Seed & LLM Expansion  \\ \midrule
		A1  & Gender--male & \tabincell{l}{The gender of this\\pedestrian is male.} & \tabincell{l}{Prioritize full body cues related to gait. The pedestrian has a\\male-presenting appearance while walking.} \\
		A5  & Hair--long & \tabincell{l}{The hair of this\\pedestrian is long.} & \tabincell{l}{Pay attention to head details during walking. This walker has hair\\extending below the head or neck.} \\
		A12 & \tabincell{c}{Upper arm\\movement--yes} & \tabincell{l}{The upper arms of this\\pedestrian move.} & \tabincell{l}{Emphasize upper body appearance and motion patterns. The upper arms\\change position clearly over the gait sequence.} \\
		A15 & Limping--yes & \tabincell{l}{The legs of this\\pedestrian are limping.} & \tabincell{l}{Focus more on lower body characteristics in walking. The pedestrian walks\\with a visible limp.} \\
		\bottomrule
	\end{tabular}}
\end{table*}

\subsubsection{Per-Attribute Temporal Aggregation}

Attributes such as arm swing direction, dragging feet, and limping can depend on body configurations observed at different points in a walking sequence.
We aggregate this evidence using multi-head attention with \(M\) learnable temporal queries \(\{\mathbf{q}_{m}\in\mathbb{R}^{d}\}_{m=1}^{M}\), one for each gait attribute.
For each spatial token position \(j\), query \(\mathbf{q}_{m}\) attends along the temporal dimension to the \(T\) frame tokens at the same token position.
We expand the attention computation below to make temporal weighting at each spatial position explicit.
For notational clarity, the query and token normalization operations preceding the projections are omitted below.
Let \(H_{\mathrm{att}}\) denote the number of attention heads, let \(h=1,\ldots,H_{\mathrm{att}}\) index a head, and let \(d_h=d/H_{\mathrm{att}}\) be the corresponding head dimension, assuming that \(H_{\mathrm{att}}\) divides \(d\).
The normalized attention weight is
\begin{equation}
	\alpha_{m,f,j}^{(h)} =
	\operatorname{softmax}_{f}
	\left(
		\tau_h
		\operatorname{sim}\!\left(
			\mathbf{W}_{Q}^{(h)}\mathbf{q}_{m},
			\mathbf{W}_{K}^{(h)}\mathbf{h}_{f,j}
		\right)
	\right).
\end{equation}
Here, \(\operatorname{softmax}_{f}\) normalizes over \(f=1,\ldots,T\) for each fixed \((m,j,h)\).
The matrices \(\mathbf{W}_{Q}^{(h)},\mathbf{W}_{K}^{(h)}\in\mathbb{R}^{d_h\times d}\) are the head-wise query and key projections, and \(\operatorname{sim}(\cdot,\cdot)\) is cosine similarity.
The positive scaling factor \(\tau_h\) is learned independently for each head through a log-space parameterization.
The attended temporal evidence is computed as
\begin{equation}
	\tilde{\mathbf{h}}_{m,j}
	=
	\mathbf{W}_{O}
	\operatorname{Concat}_{h=1}^{H_{\mathrm{att}}}
	\left(
		\sum_{f=1}^{T}
		\alpha_{m,f,j}^{(h)}
		\mathbf{W}_{V}^{(h)}\mathbf{h}_{f,j}
	\right),
\end{equation}
where \(\mathbf{W}_{V}^{(h)}\in\mathbb{R}^{d_h\times d}\) is the head-wise value projection, \(\mathbf{W}_{O}\in\mathbb{R}^{d\times d}\) is the output projection, and \(\operatorname{Concat}\) concatenates the head outputs.
The projection matrices and attention scaling factors are shared across attributes, whereas each attribute has its own temporal query.

We use the temporal mean as the residual path and add the attention output scaled by a learnable channel-wise vector:
\begin{equation}
	\bar{\mathbf{h}}_{m,j}
	=
	\frac{1}{T}\sum_{f=1}^{T}\mathbf{h}_{f,j}
	+
	\boldsymbol{\eta}\odot \tilde{\mathbf{h}}_{m,j},
\end{equation}
where \(\boldsymbol{\eta}\in\mathbb{R}^{d}\) is a learnable residual scale shared across attributes and \(\odot\) denotes element-wise multiplication.
The aggregation preserves the spatial layout and yields a separate token matrix for each attribute:
\(\bar{\mathbf{H}}_{m}=[\bar{\mathbf{h}}_{m,1},\ldots,\bar{\mathbf{h}}_{m,L}]^{\top}\in\mathbb{R}^{L\times d}\).
Compared with uniform mean pooling, learnable-query aggregation assigns attribute-specific weights to the observed frame features.

\subsubsection{Per-Attribute Spatial Aggregation}

We next pool each attribute-specific token map into a single vector using a shared multi-head attention operator with \(M\) learnable spatial queries \(\{\mathbf{p}_{m}\in\mathbb{R}^{d}\}_{m=1}^{M}\), one for each gait attribute.
For brevity, we use compact multi-head attention notation here, omitting the head-wise projection and aggregation steps.
For attribute \(a_m\), spatial query \(\mathbf{p}_m\) attends to all valid spatial tokens in \(\bar{\mathbf{H}}_m\), which serves as both keys and values:
\begin{equation}
	\mathbf{u}_m
	=
	\operatorname{MHA}
	\left(
		\mathbf{p}_m,
		\bar{\mathbf{H}}_m,
		\bar{\mathbf{H}}_m;
		\boldsymbol{\theta}_{\mathrm{attn}}
	\right)
	\in\mathbb{R}^{d},
\end{equation}
where \(\operatorname{MHA}\) denotes multi-head attention with weights normalized over valid spatial token positions, and \(\boldsymbol{\theta}_{\mathrm{attn}}\) denotes its parameters shared by all attributes.
We then apply layer normalization to \(\mathbf{u}_m\), pass the normalized features through a multilayer perceptron, and add its output to \(\mathbf{u}_m\).
Both layer normalization and the multilayer perceptron use parameters shared across attributes.
The result is \(\ell_2\)-normalized to obtain \(\mathbf{v}_m\in\mathbb{R}^{d}\), whose inner product with a normalized textual class prototype equals cosine similarity.

\subsubsection{Attribute-Specific Sequence Embeddings}

The two aggregation stages maintain a consistent attribute correspondence: spatial query \(\mathbf{p}_m\) pools the token map produced using temporal query \(\mathbf{q}_m\).
Retaining the spatial layout during temporal aggregation allows spatial pooling to operate on evidence accumulated across frames at each spatial position.
The queries are specific to each attribute, while the remaining aggregation parameters are shared across attributes within each stage.
The encoder outputs \(M\) unit-length visual embeddings \(\{\mathbf{v}_m\}_{m=1}^{M}\), each used to match the textual class prototypes of the corresponding attribute.

\subsection{Attribute Text Encoder}
\label{sec:attribute_text_encoder}

The attribute text encoder maps each predefined candidate label to a bank of semantic embeddings that can be directly compared with gait sequence embeddings.
Its textual inputs are constructed in two stages: template-based label verbalization and LLM-assisted text expansion.

First, we convert each isolated label \(\ell_m^n\) into a sentence-level template seed \(t_{m,n}^{\mathrm{seed}}\).
The template follows the slot-based form
\begin{equation}
	\begin{aligned}
		t_{m,n}^{\mathrm{seed}}={}&\text{``The \{attribute region or property\} of}\\
		&\text{this pedestrian is/are \{label state\}.''}
	\end{aligned}
\end{equation}
The two slots specify what visual property should be examined and which candidate state it should express, respectively.
We instantiate this basic form with attribute-specific nouns, copulas, and predicates to maintain grammaticality and preserve the precise annotation semantics.
For example, \textit{Gender--male} and \textit{Hair--long} are verbalized as \textit{The gender of this pedestrian is male} and \textit{The hair of this pedestrian is long}, whereas the action-oriented class \textit{Upper arm movement--yes} uses \textit{The upper arms of this pedestrian move}.
These concise seeds provide a one-to-one semantic anchor for every attribute--label pair and constrain the subsequent language expansion.

Second, we build on LLM-based class description generation~\cite{pratt2023does} to construct a richer set of complete descriptions through offline expansion.
For each class, the expansion starts from its template seed and attribute name.
We prompt Qwen3.5-Plus~\cite{qwen2026qwen35} to generate distinct descriptions of the same gait concept while preserving the label semantics specified by the seed.
These descriptions are combined with gait-aware contextual prefixes selected according to the attribute's body-region category.
The expanded descriptions are manually reviewed for consistency with the predefined labels.
We denote the final expanded description bank as
\begin{equation}
	\mathcal{T}_{m,n}=\left\{t_{m,n}^{q}\right\}_{q=1}^{Q},
\end{equation}
where \(Q\) is the number of retained descriptions for each candidate label.
Table~\ref{tab:text_examples} illustrates the two-stage construction with representative classes.
This construction is applied only on the text side; visual attribute specialization is provided separately by the attribute-specific learnable queries.

Each complete description is encoded by the frozen text encoder of the vision-language foundation model:
\begin{equation}
	\mathbf{e}_{m,n}^{q}=E_{\mathrm{text}}(t_{m,n}^{q}), \quad \mathbf{e}_{m,n}^{q}\in\mathbb{R}^{d}.
\end{equation}
The text encoder is kept frozen and the resulting embeddings are cached, which stabilizes the language space, avoids repeated encoding, and reduces the risk of overfitting the text branch to attribute supervision.
For an attribute \(a_m\) and a label \(\ell_m^n\), the output of this module is a bank of text embeddings:
\begin{equation}
	\mathcal{E}_{m,n}
	=
	\left\{
		\mathbf{e}_{m,n}^{q}
		\ \middle|\
		q=1,\ldots,Q
	\right\}.
\end{equation}
How these embeddings are sampled or aggregated to form matching scores is determined by the subsequent sequence-text matching module.

\subsection{Attribute-Centric Sequence-Text Matching}

The matching module compares the visual evidence \(\mathbf{v}_{i,m}\) of sequence \(x_i\) with textual prototypes constructed from the embedding bank \(\mathcal{E}_{m,n}\) of each candidate label \(\ell_m^n\).
Training and inference serve different roles: training uses stochastic text prototypes to learn robust alignment, while inference uses deterministic prototypes for stable attribute prediction.
Superscripts \(\mathrm{tr}\) and \(\mathrm{te}\) distinguish the training- and test-time textual prototypes, respectively.

\subsubsection{Training Objectives}

During training, the objective is computed over all annotated attributes \(a_m\in\mathcal{A}\).
For label \(\ell_m^n\), we randomly sample a non-empty subset \(\Omega_{m,n}\subseteq\mathcal{E}_{m,n}\) from the cached bank and form a stochastic training prototype:
\begin{equation}
	\bar{\mathbf{e}}_{m,n}^{\mathrm{tr}}
	=
	\operatorname{Normalize}\left(
	\frac{1}{|\Omega_{m,n}|}
		\sum_{\mathbf{e}\in\Omega_{m,n}}
		\operatorname{Normalize}(\mathbf{e})
		\right).
\end{equation}
The inner normalization gives equal angular weight to each sampled description, while the outer normalization yields a unit-length prototype.
Because \(\mathcal{E}_{m,n}\) contains multiple LLM-expanded descriptions, this stochastic prototype acts as semantic augmentation: the visual encoder is encouraged to align with the gait concept rather than a particular sentence.

Let \(\mathbf{v}_{i,m}\) be the normalized visual embedding of sequence \(x_i\) for attribute \(a_m\).
We adapt the sigmoid image-text matching objective~\cite{zhai2023sigmoid} to sequence--label pairs.
The training logit for label \(\ell_m^n\) is computed as
\begin{equation}
	s_{i,m,n}=\exp(\gamma_m) \, \mathbf{v}_{i,m}^{\top}\bar{\mathbf{e}}_{m,n}^{\mathrm{tr}}+\delta_m,
\end{equation}
where \(\gamma_m,\delta_m\in\mathbb{R}\) are the attribute-specific log-scale and bias shared by all candidate labels of attribute \(a_m\).
The class-wise logistic loss is
\begin{equation}
	\mathcal{J}^{\mathrm{sig}}_{i,m,n}=
	\operatorname{softplus}
	\left(
		-\left(2b_{i,m,n}-1\right)s_{i,m,n}
	\right).
\end{equation}
Here, \(\operatorname{softplus}(u)=\log(1+\exp(u))\).
Let \(\mathcal{B}\) denote the sample-index set of a training mini-batch.
Because each RA-GAR training sequence has complete annotations, the class-wise objective is averaged over the full mini-batch and all \mbox{candidate labels}:
\begin{equation}
	\mathcal{J}^{\mathrm{sig}}
	=
	\frac{1}{C}
	\sum_{m=1}^{M}
	\sum_{n=0}^{N_m-1}
	\frac{1}{|\mathcal{B}|}
	\sum_{i\in\mathcal{B}}
	\mathcal{J}^{\mathrm{sig}}_{i,m,n}.
\end{equation}

In addition, we use an attribute-wise softmax cross-entropy term over the candidate labels of each attribute:
\begin{equation}
	\mathcal{J}^{\mathrm{ce}}
	=
	\frac{1}{M}
	\sum_{m=1}^{M}
	\frac{1}{|\mathcal{B}|}
	\sum_{i\in\mathcal{B}}
		\operatorname{CE}\left(
			\mathbf{s}_{i,m},
			c_{i,m}
		\right).
\end{equation}
Here, \(\operatorname{CE}\) denotes softmax cross-entropy over the \(N_m\) candidate labels of attribute \(a_m\).
The class-wise sigmoid term learns an absolute decision boundary for each candidate label, consistent with the fixed-threshold inference rule, while the attribute-wise cross-entropy term encourages relative separation among the candidate labels of the same attribute.
The final training objective uses unit coefficients for both terms:
\begin{equation}
	\mathcal{J}
	=
	\mathcal{J}^{\mathrm{sig}}
	+
	\mathcal{J}^{\mathrm{ce}}.
\end{equation}

\subsubsection{Text-Guided Attribute Inference}

Inference turns the learned alignment into deterministic attribute prediction over the candidate-label sets.
For each evaluated attribute \(a_m\), we obtain the corresponding visual embedding \(\mathbf{v}_{i,m}\).
On the text side, no description is sampled.
All embeddings in \(\mathcal{E}_{m,n}\) are averaged to form the test-time prototype for label \(\ell_m^n\):
\begin{equation}
	\bar{\mathbf{e}}_{m,n}^{\mathrm{te}}
	=
	\operatorname{Normalize}\left(
	\frac{1}{|\mathcal{E}_{m,n}|}
	\sum_{\mathbf{e}\in\mathcal{E}_{m,n}}
	\operatorname{Normalize}(\mathbf{e})
	\right).
\end{equation}

The inference logit measures how well sequence \(x_i\) matches label \(\ell_m^n\):
\begin{equation}
	s_{i,m,n}^{\mathrm{te}}=\exp(\gamma_m) \, \mathbf{v}_{i,m}^{\top}\bar{\mathbf{e}}_{m,n}^{\mathrm{te}}+\delta_m.
\end{equation}
The output of text-guided inference is the score vector
\(\mathbf{s}_{i,m}^{\mathrm{te}}=[s_{i,m,0}^{\mathrm{te}},\ldots,s_{i,m,N_m-1}^{\mathrm{te}}]\in\mathbb{R}^{N_m}\)
for each evaluated attribute.
For metric computation, each candidate-label logit is independently binarized as
\begin{equation}
	\hat{b}_{i,m,n}
	=
		\mathbb{I}\!\left[s_{i,m,n}^{\mathrm{te}}>0\right].
\end{equation}
The zero-logit threshold corresponds to a sigmoid probability of 0.5 and is applied independently to every candidate label; no within-attribute argmax operation is used.
Although each ground-truth attribute is categorical, candidate-wise thresholding is retained to follow the unified binary-column evaluation protocol used by gait and pedestrian attribute recognition benchmarks~\cite{tang2019improving,song2023gait}.
The class-wise mean accuracy and instance-level metrics are then computed directly from these candidate-label predictions.

\subsection{Attribute Pre-training for Gait Recognition}

To assess the transferability of the learned gait semantics, we use the Gait-VL visual encoder as an upstream representation source for identity-oriented gait recognition.
Recent LVM-based gait recognition pipelines, such as BigGait~\cite{ye2024biggait} and BiggerGait~\cite{ye2025biggergait}, show that generic foundation-model features are useful upstream representations.
However, these upstream models are usually trained for general visual understanding rather than for walking semantics.
Gait-relevant factors such as body structure, head posture, arm swing, and lower-body motion may be present in the representation, but their correspondence to interpretable gait attributes remains implicit.

We therefore use gait attribute recognition as a sequence-level proxy task for gait representation pre-training.
The pre-training objective does not optimize identity classification or metric-learning losses.
Instead, sequence-text matching encourages the encoder to associate multi-frame walking evidence with explicit attribute semantics.
Identity labels do not enter the pre-training objective.
Propagating the attribute losses through the visual encoder encourages it to retain visual evidence of body shape, posture, and motion that may also support downstream identity recognition.

For downstream gait recognition, we adopt a direct and controlled transfer protocol.
Given a gait sequence \(x\), let \(F_{\mathrm{LVM}}\) and \(F_{\mathrm{Gait\text{-}VL}}\) denote the original generic visual encoder and the visual encoder obtained through Gait-VL attribute pre-training, respectively.
Let \(\mathcal{R}(\cdot)\) denote the unchanged downstream mapping, including the original feature processing and identity recognizer.
Both visual encoders provide the interface required by \(\mathcal{R}\), and the replacement is expressed as
\begin{equation}
	\mathcal{R}\!\left(F_{\mathrm{LVM}}(x)\right)
	\longrightarrow
	\mathcal{R}\!\left(F_{\mathrm{Gait\text{-}VL}}(x)\right).
\end{equation}
Thus, the generic upstream visual encoder is replaced by the visual encoder obtained through Gait-VL attribute pre-training, while all components represented by \(\mathcal{R}\), together with the identity training objective, gallery/probe protocol, and evaluation metric of BigGait and BiggerGait, are kept unchanged.
This protocol evaluates the transferability of attribute-supervised representations within existing gait recognition architectures.

\section{Experiments}
\label{sec:experiment}

To assess the effectiveness of Gait-VL, we first evaluate its gait attribute recognition performance on RA-GAR.
We then examine whether attribute pre-training benefits downstream gait identity recognition across pipelines and benchmarks.

\subsection{Evaluation Protocols}

\subsubsection{Gait Attribute Recognition}

We follow the RA-GAR split defined in Section~\ref{sec:ragar_split}.
The full test set is used as the primary evaluation set, and results are also reported separately for the two acquisition scenarios.
For each method, we perform separate evaluation runs on the two scenario-defined subsets and the full test set, using the same model checkpoint and inference configuration without subset-specific tuning.
All methods are evaluated on the same 15 annotated gait attributes under this split.
Within this protocol, Gait-VL (Zero-Shot) serves as the zero-shot comparison: it is an evaluation-only sequence adaptation of pretrained SigLIP2~\cite{tschannen2025siglip} without RA-GAR optimization.
It follows the temporal-before-spatial organization of Gait-VL and uses the LLM-expanded description banks encoded by the pretrained SigLIP2 text encoder; its zero-shot-specific initialization is described below.

We report the attribute-based mean accuracy (mA) and two instance-based metrics, accuracy (Acc.) and F1 score.
To handle binary and multi-class attributes in a unified manner, each candidate attribute label is represented as a binary decision column, yielding 45 columns in total.
For all compared methods, each candidate-label logit is independently thresholded at zero, which corresponds to a sigmoid probability threshold of 0.5.
Let \(\kappa=(m,n)\in\mathcal{C}\) index one candidate label in the complete candidate-label set defined in Section~\ref{sec:methods}.
Following gait and pedestrian attribute recognition work~\cite{wu2020person,song2023gait,wang2022pedestrian}, we compute mA as
\begin{equation}
	\mathrm{mA}
	=
	\frac{1}{2|\mathcal{C}|}
	\sum_{\kappa\in\mathcal{C}}
	\left(
	\frac{\mathrm{TP}_{\kappa}}{\mathrm{TP}_{\kappa}+\mathrm{FN}_{\kappa}}
	+
	\frac{\mathrm{TN}_{\kappa}}{\mathrm{TN}_{\kappa}+\mathrm{FP}_{\kappa}}
	\right),
\end{equation}
where \(\mathrm{TP}_{\kappa}\), \(\mathrm{TN}_{\kappa}\), \(\mathrm{FP}_{\kappa}\), and \(\mathrm{FN}_{\kappa}\) are the numbers of true positives, true negatives, false positives, and false negatives for candidate label \(\kappa\), respectively.

For instance-based metrics, \(\mathrm{TP}_i\), \(\mathrm{TN}_i\), \(\mathrm{FP}_i\), and \(\mathrm{FN}_i\) are accumulated over all candidate labels in \(\mathcal{C}\) for the \(i\)-th test sequence.
Let \(N_{\mathrm{test}}\) denote the number of test sequences, and define
\(\mathrm{Precision}_i=\mathrm{TP}_i/(\mathrm{TP}_i+\mathrm{FP}_i)\) and
\(\mathrm{Recall}_i=\mathrm{TP}_i/(\mathrm{TP}_i+\mathrm{FN}_i)\).
Both instance-based metrics are computed per sequence and then averaged over the test set:
\begin{equation}
	\mathrm{Acc}
	=
	\frac{1}{N_{\mathrm{test}}}
	\sum_{i=1}^{N_{\mathrm{test}}}
	\frac{\mathrm{TP}_i+\mathrm{TN}_i}
	{\mathrm{TP}_i+\mathrm{TN}_i+\mathrm{FP}_i+\mathrm{FN}_i},
\end{equation}
\begin{equation}
	\mathrm{F1}
	=
	\frac{1}{N_{\mathrm{test}}}
	\sum_{i=1}^{N_{\mathrm{test}}}
	\frac{2\,\mathrm{Precision}_i\,\mathrm{Recall}_i}
	{\mathrm{Precision}_i+\mathrm{Recall}_i}.
\end{equation}
For each sequence, precision and F1 are defined as zero when there are no true positives.

\subsubsection{Attribute Pre-training Evaluation}

We evaluate the utility of our SigLIP2-based visual encoder for identity-oriented gait recognition and the further benefit of text-guided gait attribute training.
This experiment follows the evaluation protocols of BigGait~\cite{ye2024biggait} and BiggerGait~\cite{ye2025biggergait} on CCPG~\cite{li2023in}, SUSTech1K~\cite{shen2023lidargait}, and CCGR-MINI~\cite{zou2024cross}.
We report Rank-1 accuracy as the primary metric and Rank-5 accuracy on CCGR-MINI.
Following the terminology above, Gait-VL (Zero-Shot) and Gait-VL denote the visual representations of our SigLIP2-based gait adaptation without and with RA-GAR attribute training, respectively.
Here, ``Zero-Shot'' refers to the absence of RA-GAR optimization rather than to the downstream identity-recognition training protocol.

\begin{table*}[t]
	\caption{Gait attribute recognition performance on the two scenario-defined subsets (Scenario 1 and Scenario 2) and the full RA-GAR test set (Full Set).
		``Type'' denotes the input modality.
		Gait-VL (Zero-Shot) is evaluated without RA-GAR optimization. All results are reported in percentages. The best and second-best results in each column are in \textbf{bold} and \underline{underline}, respectively.}
	\label{tab:gar_main}
	\centering
	\setlength{\tabcolsep}{8pt}
	\renewcommand{\arraystretch}{1.1}
	\resizebox{0.99\textwidth}{!}{%
		\begin{tabular}{lc|ccc|ccc|ccc}
			\toprule
			\multirow{3}{*}{Method} & \multirow{3}{*}{Type} & \multicolumn{3}{c|}{Scenario 1} & \multicolumn{3}{c|}{Scenario 2} & \multicolumn{3}{c}{Full Set} \\
			\cmidrule(lr){3-5} \cmidrule(lr){6-8} \cmidrule(lr){9-11}
			& & F1 & Acc. & mA & F1 & Acc. & mA & F1 & Acc. & mA \\
			\midrule
			GaitGraph~\cite{teepe2021gaitgraph} & \multirow{5}{*}{Pose} & 69.38 & 82.31 & 59.75 & 74.63 & 85.15 & 63.26 & 71.76 & 83.60 & 62.26 \\
			GaitGraph2~\cite{teepe2022towards} & & 72.62 & 83.81 & 63.06 & 78.29 & 86.89 & 67.79 & 75.24 & 85.24 & 66.13 \\
			GaitTR~\cite{zhang2023spatial} & & 74.19 & 83.97 & 65.22 & 79.71 & 87.29 & 70.05 & 76.72 & 85.49 & 68.33 \\
			GPGait~\cite{fu2023gpgait} & & 75.62 & 84.10 & 66.76 & 80.50 & 87.25 & 70.94 & 77.86 & 85.55 & 69.68 \\
			GPGait++~\cite{meng2025fastposegait} & & 72.54 & 82.49 & 62.15 & 75.63 & 84.48 & 65.80 & 73.96 & 83.40 & 64.75 \\
			\midrule
			GaitSet~\cite{chao2019gaitset} & \multirow{8}{*}{Silhouette} & 74.89 & 83.53 & 64.81 & 77.62 & 85.45 & 65.84 & 76.14 & 84.41 & 66.55 \\
			GaitPart~\cite{fan2020gaitpart} & & 75.72 & 84.23 & 65.84 & 79.36 & 86.66 & 67.79 & 77.39 & 85.35 & 67.96 \\
			GaitGL~\cite{lin2021gait} & & \underline{77.91} & \underline{85.76} & 65.71 & 82.73 & \textbf{88.93} & 69.86 & \underline{80.12} & \underline{87.21} & 68.83 \\
			DeepGaitV2-2D~\cite{fan2023exploring} & & 76.73 & 84.81 & 64.07 & 80.64 & 87.45 & 65.88 & 78.52 & 86.02 & 65.92 \\
			DeepGaitV2-3D~\cite{fan2023exploring} & & 75.43 & 84.07 & 63.06 & 78.58 & 86.35 & 64.36 & 76.87 & 85.12 & 64.72 \\
			DeepGaitV2-P3D~\cite{fan2023exploring} & & 75.33 & 84.05 & 62.20 & 78.82 & 86.47 & 63.24 & 76.93 & 85.16 & 63.58 \\
			GAR-Net~\cite{song2023gait} & & 76.82 & 83.99 & 64.23 & 80.72 & 86.98 & 67.12 & 78.61 & 85.36 & 66.64 \\
			CLIP-GAR~\cite{wang2025ra} & & 76.59 & 84.45 & \underline{67.10} & \underline{83.31} & \underline{88.92} & \underline{73.09} & 79.67 & 86.50 & \underline{71.19} \\
			\midrule
			ALM~\cite{tang2019improving} & \multirow{4}{*}{RGB} & 74.77 & 82.76 & 64.53 & 65.29 & 75.96 & 55.32 & 70.46 & 79.67 & 60.44 \\
			A-Solution~\cite{zhou2023solution} & & 75.29 & 83.85 & 64.12 & 67.12 & 79.08 & 57.56 & 71.54 & 81.66 & 61.67 \\
			Gait-VL (Zero-Shot) & & 45.10 & 63.40 & 54.83 & 43.64 & 62.43 & 52.45 & 44.43 & 62.96 & 54.67 \\
			\bftab{Gait-VL (Ours)} & & \textbf{80.34} & \textbf{86.46} & \textbf{70.15} & \textbf{83.82} & \textbf{88.93} & \textbf{73.55} & \textbf{81.94} & \textbf{87.59} & \textbf{73.28} \\
			\bottomrule
	\end{tabular}}
\end{table*}

\subsection{Implementation Details}

We use SigLIP2-Base-Patch16-NaFlex as the foundation model and initialize it from the official pretrained checkpoint.
For Gait-VL (Zero-Shot), we set the temporal residual scale \(\boldsymbol{\eta}\) to zero, initialize all attribute-specific spatial queries from the same pretrained global pooling query, and initialize every attribute-specific log-scale and bias by copying the corresponding pretrained SigLIP2 values.
This configuration is used without RA-GAR optimization.
When training Gait-VL, each element of \(\boldsymbol{\eta}\) is initialized to \(1\times10^{-3}\), and the temporal attention uses 16 heads, with a learnable scaling factor initialized to 64.0 for each head.
For its spatial pooling, the shared parameters of the attention operator, layer normalization, and multilayer perceptron are initialized from the backbone checkpoint, while each attribute-specific spatial query is initialized by perturbing the pretrained global pooling query.
The attribute-specific log-scales and biases are likewise initialized by copying the pretrained SigLIP2 values and are then optimized separately during training.
Each RGB frame is resized to a height\(\times\)width resolution of \(448\times224\).
Training mini-batches use 16--48 randomly sampled frames per sequence.
At inference, we use all available frames up to a maximum of 512 per sequence.
We use \(Q=30\) descriptions per class, randomly sampling \(1\)--\(Q\) descriptions during training and using all \(Q\) at inference.

The text encoder remains frozen throughout training.
The vision backbone is frozen for the first 10k iterations and then unfrozen, while the temporal aggregation module, spatial pooling module, and per-attribute logit parameters are optimized throughout training.
We train the model for 40k iterations using AdamW.
The learning rate is \(1\times10^{-5}\) for the backbone and \(1\times10^{-4}\) for the temporal aggregation, spatial pooling, and logit parameters.
Weight decay is set to 0, and gradient norm is clipped to 5.0.
We use a warm-up multi-step learning rate schedule with 5k warm-up iterations, a warm-up factor of 0.1, and a milestone at 30k iterations.
The global mini-batch sampler is configured with 2 identities and 8 sequences per identity.
Training is performed with automatic mixed precision on 8 A100 40GB GPUs.

We use all 15 annotated attributes and include both natural- and instructed-state sequences, so that the experiments cover the complete attribute taxonomy and the collected walking variations.
For gait attribute recognition, we use only the official training split of RA-GAR, keeping all test subjects unseen.
For downstream identity recognition, attribute pre-training uses the entire RA-GAR dataset as an upstream semantic corpus, without identity classification or metric-learning objectives.
The resulting checkpoint is then evaluated as Gait-VL in BigGait and BiggerGait.
It is worth emphasizing that RA-GAR shares no subjects with existing gait recognition benchmarks such as CCPG, SUSTech1K, or CCGR-MINI.

To support reproducibility, the Gait-VL code and experimental configurations will be made available at \url{https://github.com/BNU-IVC/RA-GAR}.
RA-GAR is distributed under controlled access, and the repository will provide instructions for requesting the dataset and its associated terms of use.

\begin{table}[t]
	\caption{Per-attribute mA of GAR-Net, CLIP-GAR, and Gait-VL on the full RA-GAR test set.
	Each method uses the same checkpoint and inference settings as in Table~\ref{tab:gar_main}.
	For each attribute, mA (\%) is averaged over its candidate labels.
	Attributes are grouped by body region. The best and second-best results in each row are in \textbf{bold} and \underline{underline}.}
	\label{tab:gar_subset}
	\centering
	\normalsize
	\setlength{\tabcolsep}{3pt}
	\renewcommand{\arraystretch}{1.1}
	\resizebox{0.99\columnwidth}{!}{%
	\begin{tabular}{cl|ccc}
		\toprule
		Region & Attribute & \tabincell{c}{GAR-Net\\\cite{song2023gait}} & \tabincell{c}{CLIP-GAR\\\cite{wang2025ra}} & \bftab{\tabincell{c}{Gait-VL\\(Ours)}} \\
		\midrule
		\multirow{4}{*}{\tabincell{c}{Full\\body}}
		& A1 Gender             & 92.76 & \underline{94.20} & \textbf{96.73} \\
		& A2 Age                & 79.25 & \underline{80.04} & \textbf{81.90} \\
		& A3 Height             & \underline{61.04} & 60.10 & \textbf{61.47} \\
		& A4 Body mass          & 52.76 & \textbf{54.59} & \underline{53.05} \\
		\midrule
		\multirow{2}{*}{Head}
		& A5 Hair                  & 68.71 & \underline{73.14} & \textbf{77.48} \\
		& A6 Forward head position & 70.33 & \underline{73.81} & \textbf{76.81} \\
		\midrule
		\multirow{6}{*}{\tabincell{c}{Upper\\body}}
		& A7 Uneven shoulders       & 60.98 & \underline{80.13} & \textbf{80.51} \\
		& A8 Humpback               & 80.39 & \underline{86.16} & \textbf{87.35} \\
		& A9 Hands                  & 51.38 & \underline{53.34} & \textbf{60.70} \\
		& A10 Arm swing direction   & 61.10 & \underline{64.25} & \textbf{67.18} \\
		& A11 Arm swing amplitude & 64.70 & \underline{70.41} & \textbf{73.42} \\
		& A12 Upper arm movement    & 63.70 & \underline{70.28} & \textbf{75.70} \\
		\midrule
		\multirow{3}{*}{\tabincell{c}{Lower\\body}}
		& A13 Toe                & 79.95 & \textbf{85.37} & \underline{84.84} \\
		& A14 Drag feet          & 60.47 & \underline{75.84} & \textbf{77.11} \\
		& A15 Limping            & 85.75 & \textbf{89.78} & \underline{86.68} \\
		\bottomrule
	\end{tabular}}
\end{table}

\subsection{Performance Comparison}

\subsubsection{Gait Attribute Recognition}

We first compare performance on the full RA-GAR test set using Table~\ref{tab:gar_main}.
The compared methods are grouped into pose-based, silhouette-based, and RGB-based approaches.
The RGB baselines include two representative pedestrian attribute recognition methods: ALM~\cite{tang2019improving}, which uses weakly supervised multi-scale attribute-specific localization, and A-Solution~\cite{zhou2023solution}, which uses attribute-disentangled learning to mitigate attribute co-occurrence bias.
We additionally include Gait-VL (Zero-Shot), our sequence adaptation of pretrained SigLIP2 without RA-GAR optimization, as the zero-shot vision-language baseline.

Gait-VL achieves the highest full-set F1, Acc., and mA among the compared methods, with an mA gain of 2.09 percentage points over CLIP-GAR~\cite{wang2025ra}, the method proposed in our conference version and the strongest baseline on this metric.
Its gains over RGB attribute recognition baselines and the zero-shot vision-language baseline further support the effectiveness of the complete framework for learning fine-grained gait attributes from RGB sequences.

To examine how the overall advantage is distributed across attributes, Table~\ref{tab:gar_subset} compares Gait-VL with two representative prior gait attribute recognition methods.
We select GAR-Net as the model introduced with the earlier MA-Gait benchmark and CLIP-GAR as the text-guided predecessor from our conference version.
Gait-VL achieves the best result on 12 of the 15 attributes.
Relative to CLIP-GAR, the gains on \textit{Hands} and \textit{Upper arm movement} are 7.36 and 5.42 percentage points, respectively, showing that the overall improvement includes local posture and motion attributes.
However, CLIP-GAR remains stronger on \textit{Body mass}, \textit{Toe}, and \textit{Limping}, indicating that the benefits vary across semantic categories.

\begin{table*}[t]
	\caption{Evaluation of the proposed Gait-VL visual encoders in existing gait recognition pipelines.
		Gait-VL (Zero-Shot) and Gait-VL denote the SigLIP2-based upstream representations without and with RA-GAR attribute training, respectively.
		Within each pipeline, the downstream recognition architecture, training objectives, and evaluation protocols are kept unchanged.
		Input resolution is reported as height$\times$width. CL, UP, DN, and BG on CCPG denote clothes-changing, upper-body, lower-body, and bag-carrying protocols, respectively.
		Cloth. denotes the clothes-changing condition on SUSTech1K.
		All columns report Rank-1 accuracy (\%) except R-5, which reports Rank-5 accuracy (\%). The best and second-best results in each column within each pipeline are in \textbf{bold} and \underline{underline}, respectively.}
	\label{tab:lvm_gait_transfer}
	\centering
	\setlength{\tabcolsep}{8pt}
	\renewcommand{\arraystretch}{1.1}
	\resizebox{0.99\textwidth}{!}{%
		\begin{tabular}{l|l|c|cccc|cc|cc}
			\toprule
			\multirow{2}{*}{Pipeline}
			& \multirow{2}{*}{Upstream}
			& \multirow{2}{*}{\tabincell{c}{Input\\Resolution}}
			& \multicolumn{4}{c|}{CCPG~\cite{li2023in}}
			& \multicolumn{2}{c|}{SUSTech1K~\cite{shen2023lidargait}}
			& \multicolumn{2}{c}{CCGR-MINI~\cite{zou2024cross}} \\
			\cmidrule(lr){4-7} \cmidrule(lr){8-9} \cmidrule(lr){10-11}
			& & & CL & UP & DN & BG
			& Cloth. & R-1
			& R-1 & R-5 \\
			\midrule
			\multirow{3}{*}{BigGait~\cite{ye2024biggait}}
			& DINOv2-S~\cite{oquab2023dinov2} & $448\times224$ & 82.6 & 85.9 & 87.1 & 93.1 & 43.7 & 56.4 & 7.4 & 16.3 \\
			& Gait-VL (Zero-Shot) & $448\times224$ & \underline{90.9} & \underline{91.9} & \underline{93.2} & \underline{95.4} & \underline{47.9} & \underline{63.1} & \underline{10.5} & \underline{20.9} \\
			& \bftab{Gait-VL (Ours)} & $448\times224$ & \textbf{92.2} & \textbf{92.6} & \textbf{93.5} & \textbf{95.8} & \textbf{53.7} & \textbf{65.0} & \textbf{11.1} & \textbf{22.3} \\
			\midrule
			\multirow{5}{*}{BiggerGait~\cite{ye2025biggergait}}
			& SAM-S~\cite{kirillov2023segment} & $512\times256$ & 86.9 & 89.4 & 92.3 & 95.8 & 60.8 & 74.4 & 9.1 & 17.2 \\
			& CLIP-S~\cite{radford2021learning} & $224\times224$ & 78.9 & 83.8 & 87.9 & 96.1 & 63.1 & 81.5 & 13.9 & 24.2 \\
			& DINOv2-S~\cite{oquab2023dinov2} & $448\times224$ & 89.0 & 91.9 & 94.0 & \underline{97.2} & 69.5 & 80.4 & 14.5 & 25.3 \\
			& Gait-VL (Zero-Shot) & $448\times224$ & \underline{94.3} & \underline{94.9} & \underline{96.1} & 96.9 & \underline{72.7} & \underline{85.4} & \underline{17.7} & \underline{31.1} \\
			& \bftab{Gait-VL (Ours)} & $448\times224$ & \textbf{95.5} & \textbf{96.2} & \textbf{96.7} & \textbf{97.5} & \textbf{73.6} & \textbf{86.8} & \textbf{22.7} & \textbf{36.2} \\
			\bottomrule
	\end{tabular}}
\end{table*}

\subsubsection{Performance by Acquisition Scenario}

To examine how the compared methods behave under different acquisition configurations, we report their performance separately on the two scenario-defined subsets in Table~\ref{tab:gar_main}.
Gait-VL achieves the highest F1 and mA in both scenarios, as well as the highest Acc. in Scenario 1, and ties GaitGL for the highest Acc. in Scenario 2.
Its advantage over the best-performing baseline for each metric is more pronounced in Scenario 1, whereas the performance gap becomes smaller in Scenario 2.
These results show that the overall advantage of Gait-VL is maintained across the two subsets, although its magnitude varies with the acquisition setting.

Comparing the two scenarios also reveals a clear difference across method families.
All pose- and silhouette-based gait methods achieve higher scores on the three metrics in Scenario 2, and Gait-VL follows the same trend.
By contrast, the generic RGB attribute recognition methods and Gait-VL (Zero-Shot) obtain lower scores in Scenario 2 than in Scenario 1.
Scenario 2 is therefore not uniformly easier for all visual models; rather, its acquisition setting appears particularly favorable to methods built around gait-specific representations of body structure, posture, and motion.

The two scenarios differ in camera height, view distribution, acquisition environment, and participant composition.
In particular, Scenario 2 uses cameras mounted at 2.5 meters instead of 1.3 meters.
The consistent improvement of gait-specific methods suggests that this higher camera placement may be useful for fine-grained gait attribute analysis, although its effect cannot be completely isolated from the other differences between the two scenarios.

\subsubsection{Attribute Pre-training Evaluation}

Beyond attribute recognition, Table~\ref{tab:lvm_gait_transfer} examines the utility of our gait representations for identity recognition in BigGait and BiggerGait on CCPG, SUSTech1K, and CCGR-MINI.
Following the evaluation protocol above, we first compare Gait-VL (Zero-Shot) with generic upstream models, then assess the additional gains from RA-GAR attribute training.

Specifically, Gait-VL (Zero-Shot) outperforms DINOv2-S~\cite{oquab2023dinov2} on all reported protocols in BigGait.
In BiggerGait, it also surpasses SAM-S~\cite{kirillov2023segment}, CLIP-S~\cite{radford2021learning}, and DINOv2-S~\cite{oquab2023dinov2} on all reported protocols except CCPG-BG.
Building on this representation, Gait-VL improves on Gait-VL (Zero-Shot) across all reported protocols in both downstream pipelines.
Notably, in BigGait, Gait-VL improves Rank-1 accuracy over Gait-VL (Zero-Shot) by 5.8 percentage points under the clothes-changing condition of SUSTech1K.
In BiggerGait, the corresponding gain on CCGR-MINI is 5.0 percentage points.
It is worth emphasizing that these gains are obtained without any changes to the downstream model architectures or hyperparameters.
Together with the attribute recognition results, they support gait attributes as both prediction targets and transferable supervision for identity recognition.

\subsection{Ablation Studies}

\subsubsection{Effect of Input Resolution and Model Scale}

Table~\ref{tab:ablation_backbone}(a, b) evaluates the effect of input resolution for the evaluation-only Gait-VL (Zero-Shot) configuration and the RA-GAR-trained Gait-VL, respectively, with SigLIP2-Base-Patch16-NaFlex fixed.
At matched heights of 256 and 448, portrait inputs outperform square inputs on all three metrics for both settings.
The \(448\times224\) setting achieves the best or tied-best scores on each metric for both settings, while increasing the portrait resolution to \(512\times256\) yields no further improvement.
These results support the portrait input configuration used in the main experiments, consistent with the aspect-ratio preservation criterion for backbone selection.

With the input fixed at \(448\times224\), Table~\ref{tab:ablation_backbone}(c) compares the SigLIP2 Base and SO400M backbones in Gait-VL.
The larger model yields a slightly higher Acc., but neither F1 nor mA improves.
Scaling the backbone therefore offers no consistent advantage in this setting.

\begin{table}[t]
	\caption{Ablation on input resolution and model scale on the full RA-GAR test set.
	\textnormal{(a, b)} Different input resolutions (height$\times$width) with SigLIP2-Base-Patch16-NaFlex for the evaluation-only Gait-VL (Zero-Shot) configuration and the RA-GAR-trained Gait-VL, respectively.
	\textnormal{(c)} Different backbone scales for Gait-VL at the $448\times224$ main setting.
	All results are reported in percentages. The best results in each column within each block, including ties at the reported precision, are in \textbf{bold}.}
	\label{tab:ablation_backbone}
	\centering
	\setlength{\tabcolsep}{6pt}
	\renewcommand{\arraystretch}{1.1}
	\resizebox{0.99\columnwidth}{!}{%
	\begin{tabular}{l|ccc}
		\toprule
		Setting & F1 & Acc. & mA \\
		\midrule
		\multicolumn{4}{l}{\textit{(a) Input Resolution: Gait-VL (Zero-Shot)}} \\
		$256\times256$ (square)                & 39.91 & 59.94 & 53.51 \\
		$256\times128$ (portrait)              & 41.60 & 61.07 & 53.78 \\
		$448\times448$ (square)                & 42.41 & 61.61 & 53.71 \\
		$448\times224$ (portrait, main setting) & \textbf{44.43} & \textbf{62.96} & \textbf{54.67} \\
		$512\times256$ (portrait)              & 43.76 & 62.50 & 54.55 \\
		\midrule
		\multicolumn{4}{l}{\textit{(b) Input Resolution: Gait-VL}} \\
		$256\times256$ (square)                & 80.63 & 86.69 & 71.32 \\
		$256\times128$ (portrait)              & 81.02 & 86.83 & 71.91 \\
		$448\times448$ (square)                & 81.15 & 87.05 & 72.39 \\
		$448\times224$ (portrait, main setting) & \textbf{81.94} & \textbf{87.59} & \textbf{73.28} \\
		$512\times256$ (portrait)              & 81.87 & 87.55 & \textbf{73.28} \\
		\midrule
		\multicolumn{4}{l}{\textit{(c) Model Scale: Gait-VL}} \\
		SigLIP2-Base-Patch16-NaFlex~\cite{tschannen2025siglip}  & \textbf{81.94} & 87.59 & \textbf{73.28} \\
		SigLIP2-SO400M-Patch16-NaFlex~\cite{tschannen2025siglip} & 81.86 & \textbf{87.70} & 73.17 \\
		\bottomrule
	\end{tabular}}
\end{table}

\subsubsection{Effect of Attribute-Aware Sequence Encoding}

We first compare uniform mean aggregation, shared queries, and attribute-specific queries under the temporal-before-spatial order in Table~\ref{tab:ablation_seq}.
With temporal mean aggregation fixed, replacing spatial mean pooling with a shared query improves all three metrics, and attribute-specific spatial queries provide further gains.
Adding a shared temporal query does not improve all metrics, whereas using attribute-specific queries for both stages produces the best results.
The full configuration improves mA by 3.13 percentage points over uniform mean aggregation in both dimensions.

Keeping attribute-specific queries in both stages, we then reverse the aggregation order.
Temporal-before-spatial aggregation outperforms temporal-after-spatial aggregation on all three metrics.
This result supports aggregating multi-frame evidence while preserving the spatial token layout before extracting per-attribute embeddings through spatial pooling.

\subsubsection{Effect of Training Objectives}

Table~\ref{tab:ablation_loss} compares class-wise sigmoid loss, attribute-wise softmax cross-entropy, and their combination.
Sigmoid loss alone provides strong performance, whereas softmax cross-entropy alone performs substantially worse, particularly in F1.
Combining the two objectives improves all three metrics over sigmoid-only training, yielding the best results.
These results support combining candidate-wise calibration with within-attribute class discrimination in the joint objective.

\subsubsection{Effect of Attribute Text Construction}

Table~\ref{tab:ablation_text} compares three levels of attribute text construction while keeping the visual modules and training objectives unchanged.
\emph{Attribute--Label Only} uses concise attribute--label strings; \emph{Template Only} uses the sentence-level seeds defined in Section~\ref{sec:attribute_text_encoder}; and \emph{LLM Expansion} uses the description banks, incorporating both linguistic diversity and context about relevant body regions.

Moving from attribute--label strings to sentence-level templates improves all three metrics, and the expanded description banks yield further gains.
Relative to \emph{Attribute--Label Only}, the full setting improves mA by 2.74 percentage points.
These results support using complete gait-aware descriptions as semantic targets for attribute matching.

\begin{table}[t]
	\caption{Ablation on attribute-aware sequence encoding on the full RA-GAR test set.
	``Spatial Aggregation'' and ``Temporal Aggregation'' use uniform mean, a query shared across attributes, or per-attribute queries; ``Agg Mode'' denotes whether temporal aggregation is applied before or after spatial aggregation.
	All results are reported in percentages. The best results in each column are in \textbf{bold}.}
	\label{tab:ablation_seq}
	\centering
	\setlength{\tabcolsep}{5pt}
	\renewcommand{\arraystretch}{1.1}
	\resizebox{0.99\columnwidth}{!}{%
	\begin{tabular}{ccc|ccc}
		\toprule
		\tabincell{c}{Agg\\Mode} & \tabincell{c}{Spatial\\Aggregation} & \tabincell{c}{Temporal\\Aggregation} & F1 & Acc. & mA \\
		\midrule
		\multirow{5}{*}{\tabincell{c}{\emph{Temporal}\\ \emph{Before}\\ \emph{Spatial}}} & Uniform Mean & Uniform Mean & 80.47 & 86.81 & 70.15 \\
		                  & Shared Query  & Uniform Mean    & 81.15 & 87.10 & 72.06 \\
		                  & Per-Attr Query & Uniform Mean & 81.43 & 87.13 & 72.60 \\
		                  & Per-Attr Query & Shared Query & 81.45 & 87.20 & 72.51 \\
		                  & Per-Attr Query & Per-Attr Query & \textbf{81.94} & \textbf{87.59} & \textbf{73.28} \\
		\midrule
		\tabincell{c}{\emph{Temporal}\\ \emph{After}\\ \emph{Spatial}} & Per-Attr Query & Per-Attr Query & 81.13 & 87.41 & 72.53 \\
		\bottomrule
	\end{tabular}}
\end{table}

\subsubsection{Computational Cost Analysis}

We compare the computational cost of the Gait-VL visual encoder with the original SigLIP2 visual encoder at an input resolution of \(448\times224\).
Both parameter counts and FLOPs are computed exclusively for the visual pathway, with FLOPs measured per frame.
The original SigLIP2 visual encoder has 92.93M parameters and requires 72.75 GFLOPs, while Gait-VL has 95.32M parameters and requires 74.36 GFLOPs.
These correspond to increases of approximately 2.6\% in parameter count and 2.2\% in FLOPs per frame, indicating that attribute-aware temporal and spatial aggregation adds a modest overhead.

\begin{table}[t]
	\caption{Ablation on training objectives on the full RA-GAR test set.
		``Sigmoid'' denotes class-wise sigmoid loss; ``Softmax'' denotes attribute-wise cross-entropy.
		All results are reported in percentages. The best results in each column are in \textbf{bold}.}
	\label{tab:ablation_loss}
	\centering
	\setlength{\tabcolsep}{6pt}
	\renewcommand{\arraystretch}{1.1}
	\resizebox{0.8\columnwidth}{!}{%
		\begin{tabular}{cc|ccc}
			\toprule
			Sigmoid Loss & Softmax Loss & F1 & Acc. & mA \\
			\midrule
			\checkmark & --         & 81.21 & 87.39 & 71.96 \\
			--         & \checkmark & 18.05 & 67.68 & 52.76 \\
			\checkmark & \checkmark & \textbf{81.94} & \textbf{87.59} & \textbf{73.28} \\
			\bottomrule
	\end{tabular}}
\end{table}

\section{Discussion and Future Work}
\label{sec:discussion}

\subsection{Attribute-Level Gait Understanding}

Recognizing attributes of body structure, posture, and motion makes fine-grained gait understanding concrete: individual aspects of a walking sequence can be described and evaluated separately.
RA-GAR provides unified definitions, annotations, and evaluation protocols for these semantic targets.
This shared foundation supports systematic study of how reliably gait attributes can be recognized across the acquisition conditions and walking variations represented in the benchmark.

The semantic nature of these targets motivates using language to represent candidate classes directly.
By aligning attribute-specific visual representations with textual class prototypes, Gait-VL incorporates human-readable class definitions into both representation learning and prediction.
The attribute recognition results, together with the ablations on sequence encoding and text construction, support the effectiveness of matching attribute-specific walking evidence with \mbox{gait-aware descriptions}.

\subsection{Attribute Semantics as a Bridge to Gait Recognition}

Beyond their role as prediction targets, gait attributes can provide supervision for learning transferable visual representations.
Learning static attributes encourages attention to body characteristics and posture, while learning motion attributes requires evidence from walking patterns such as arm swing and foot movement.
These cues can also be relevant to identity discrimination, making attribute learning a useful pre-training task for identity-oriented gait recognition.

The transfer results in Table~\ref{tab:lvm_gait_transfer} support this connection.
Compared with representations from the same pretrained backbone without RA-GAR attribute training, Gait-VL improves performance across all reported protocols in both downstream gait recognition pipelines.
These gains are obtained without identity classification or metric-learning objectives during attribute pre-training, supporting the value of attribute supervision for downstream identity recognition.
Gait attributes therefore serve both as targets for fine-grained understanding and as transferable supervision for gait representation learning.

\subsection{Practical Implications and Responsible Use}

The attribute predictions provide a semantic interface for organizing and examining walking sequences.
For example, toe orientation and arm-swing patterns could serve as search criteria for retrieving relevant clips, selecting samples for comparative analysis, and producing concise attribute summaries for human review.
They could also help organize recognition errors by interpretable gait characteristics, supporting more targeted analysis of model behavior.
These applications are potential extensions of the present recognition framework and require direct evaluation in their intended settings.

Responsible use requires attention to prediction reliability and the handling of personal information.
The per-attribute results in Table~\ref{tab:gar_subset} show that recognition performance varies across attributes, motivating validation of the relevant attributes under the intended acquisition conditions.
Attribute summaries should be treated as model estimates and reviewed alongside the original sequences when used to interpret individual cases.
Use of RA-GAR and its derived outputs should follow the informed-consent scope, approved research purposes, and controlled-access terms of the dataset.

\begin{table}[t]
	\caption{Ablation on attribute text construction on the full RA-GAR test set.
		Attribute--Label Only uses concise attribute--label strings, while Template Only uses sentence-level template seeds.
		LLM Expansion applies the complete expansion process, including semantic diversification and attribute-relevant gait context.
		All results are reported in percentages. The best results in each column are in \textbf{bold}.}
	\label{tab:ablation_text}
	\centering
	\setlength{\tabcolsep}{10pt}
	\renewcommand{\arraystretch}{1.1}
	\resizebox{0.8\columnwidth}{!}{%
		\begin{tabular}{l|ccc}
			\toprule
			Text Construction & F1 & Acc. & mA \\
			\midrule
			Attribute--Label Only & 80.37 & 86.28 & 70.54 \\
			Template Only & 81.21 & 86.90 & 71.77 \\
			LLM Expansion & \textbf{81.94} & \textbf{87.59} & \textbf{73.28} \\
			\bottomrule
	\end{tabular}}
\end{table}

\subsection{Limitations and Future Directions}

RA-GAR brings static and motion attributes, multiple data modalities, and varied acquisition conditions into a shared annotation framework.
The current evaluation covers the attributes and collection settings represented in this benchmark.
Future extensions can broaden the range of naturally occurring walking patterns, environments, and participant populations.
The established semantic definitions and evaluation framework provide a basis for measuring progress as this coverage expands.
On the modeling side, textual descriptions provide class prototypes, while visual aggregation uses attribute-specific learnable queries.
Future work can explore text-conditioned visual queries that use attribute descriptions to guide the selection of relevant body regions and visual cues.
Evaluating such extensions on additional gait attributes would help assess whether this closer connection between language and visual aggregation supports a broader range of gait semantics.

Together, these directions would extend fine-grained gait understanding to more diverse acquisition conditions and a broader range of semantic attributes, building on the benchmark and modeling framework established in this work.

\section{Conclusion}
\label{sec:Conclusion}

This work advances fine-grained gait understanding by treating human-interpretable attributes as both recognition targets and supervision for transferable representation learning.
RA-GAR provides a large-scale multimodal benchmark for systematically studying body structure, posture, and motion through rich static and motion attribute annotations.
Gait-VL formulates attribute recognition as sequence-text matching, aligning attribute-specific multi-frame visual evidence with textual class prototypes constructed from LLM-expanded descriptions and context about relevant body regions.

Gait-VL achieves state-of-the-art attribute recognition performance on RA-GAR.
Beyond attribute prediction, attribute pre-training consistently improves identity recognition across the evaluated downstream pipelines and benchmarks.
These gains are obtained without identity classification or metric-learning objectives during attribute pre-training.
Together, these results demonstrate that learning human-interpretable gait semantics can also strengthen representations for identity recognition, connecting fine-grained gait understanding with discriminative representation learning.



\begin{thebibliography}{10}
\providecommand{\url}[1]{#1}
\csname url@samestyle\endcsname
\providecommand{\newblock}{\relax}
\providecommand{\bibinfo}[2]{#2}
\providecommand{\BIBentrySTDinterwordspacing}{\spaceskip=0pt\relax}
\providecommand{\BIBentryALTinterwordstretchfactor}{4}
\providecommand{\BIBentryALTinterwordspacing}{\spaceskip=\fontdimen2\font plus
\BIBentryALTinterwordstretchfactor\fontdimen3\font minus
  \fontdimen4\font\relax}
\providecommand{\BIBforeignlanguage}[2]{{%
\expandafter\ifx\csname l@#1\endcsname\relax
\typeout{** WARNING: IEEEtran.bst: No hyphenation pattern has been}%
\typeout{** loaded for the language `#1'. Using the pattern for}%
\typeout{** the default language instead.}%
\else
\language=\csname l@#1\endcsname
\fi
#2}}
\providecommand{\BIBdecl}{\relax}
\BIBdecl

\bibitem{bouchrika2011using}
I.~Bouchrika, M.~Goffredo, J.~Carter, and M.~Nixon, ``On using gait in forensic
  biometrics,'' \emph{Journal of Forensic Sciences}, vol.~56, no.~4, pp.
  882--889, 2011.

\bibitem{takemura2018multi}
N.~Takemura, Y.~Makihara, D.~Muramatsu, T.~Echigo, and Y.~Yagi, ``Multi-view
  large population gait dataset and its performance evaluation for cross-view
  gait recognition,'' \emph{IPSJ Transactions on Computer Vision and
  Applications}, vol.~10, pp. 1--14, 2018.

\bibitem{zhu2021gait}
Z.~Zhu, X.~Guo, T.~Yang, J.~Huang, J.~Deng, G.~Huang, D.~Du, J.~Lu, and
  J.~Zhou, ``Gait recognition in the wild: A benchmark,'' in \emph{ICCV}, 2021,
  pp. 14\,789--14\,799.

\bibitem{yu2006framework}
S.~Yu, D.~Tan, and T.~Tan, ``A framework for evaluating the effect of view
  angle, clothing and carrying condition on gait recognition,'' in \emph{ICPR},
  vol.~4, 2006, pp. 441--444.

\bibitem{zheng2022gait}
J.~Zheng, X.~Liu, W.~Liu, L.~He, C.~Yan, and T.~Mei, ``Gait recognition in the
  wild with dense {3D} representations and a benchmark,'' in \emph{CVPR}, 2022,
  pp. 20\,228--20\,237.

\bibitem{song2022casia}
C.~Song, Y.~Huang, W.~Wang, and L.~Wang, ``{CASIA-E}: a large comprehensive
  dataset for gait recognition,'' \emph{IEEE Transactions on Pattern Analysis
  and Machine Intelligence}, vol.~45, no.~3, pp. 2801--2815, 2023.

\bibitem{song2023gait}
X.~Song, S.~Hou, Y.~Huang, C.~Cao, X.~Liu, Y.~Huang, and C.~Shan, ``Gait
  attribute recognition: A new benchmark for learning richer attributes from
  human gait patterns,'' \emph{IEEE Transactions on Information Forensics and
  Security}, vol.~19, pp. 1--14, 2024.

\bibitem{makihara2011gait}
Y.~Makihara, M.~Okumura, H.~Iwama, and Y.~Yagi, ``Gait-based age estimation
  using a whole-generation gait database,'' in \emph{International Joint
  Conference on Biometrics}, 2011, pp. 1--6.

\bibitem{xu2017isir}
C.~Xu, Y.~Makihara, G.~Ogi, X.~Li, Y.~Yagi, and J.~Lu, ``The {OU-ISIR} gait
  database comprising the large population dataset with age and performance
  evaluation of age estimation,'' \emph{IPSJ Transactions on Computer Vision
  and Applications}, vol.~9, no.~1, pp. 1--14, 2017.

\bibitem{ortells2018vision}
J.~Ortells, M.~T. Herrero-Ezquerro, and R.~A. Mollineda, ``Vision-based gait
  impairment analysis for aided diagnosis,'' \emph{Medical \& biological
  engineering \& computing}, vol.~56, no.~9, pp. 1553--1564, 2018.

\bibitem{ye2024biggait}
D.~Ye, C.~Fan, J.~Ma, X.~Liu, and S.~Yu, ``{BigGait}: Learning gait
  representation you want by large vision models,'' in \emph{CVPR}, 2024, pp.
  200--210.

\bibitem{ye2025biggergait}
D.~Ye, C.~Fan, Z.~Huang, C.~Luo, J.~Li, S.~Yu, and X.~Liu, ``{BiggerGait}:
  Unlocking gait recognition with layer-wise representations from large vision
  models,'' in \emph{NeurIPS}, 2025, pp. 81\,798--81\,819.

\bibitem{wang2025ra}
C.~Wang, S.~Hou, A.~Li, Q.~Cai, and Y.~Huang, ``{RA-GAR}: A richly annotated
  benchmark for gait attribute recognition,'' in \emph{AAAI}, vol.~39, no.~7,
  2025, pp. 7591--7599.

\bibitem{li2023in}
W.~Li, S.~Hou, C.~Zhang, C.~Cao, X.~Liu, Y.~Huang, and Y.~Zhao, ``An in-depth
  exploration of person re-identification and gait recognition in
  cloth-changing conditions,'' in \emph{CVPR}, 2023, pp. 13\,824--13\,833.

\bibitem{shen2023lidargait}
C.~Shen, C.~Fan, W.~Wu, R.~Wang, G.~Q. Huang, and S.~Yu, ``{LiDARGait}:
  Benchmarking {3D} gait recognition with point clouds,'' in \emph{CVPR}, 2023,
  pp. 1054--1063.

\bibitem{zou2024cross}
S.~Zou, C.~Fan, J.~Xiong, C.~Shen, S.~Yu, and J.~Tang, ``Cross-covariate gait
  recognition: A benchmark,'' in \emph{AAAI}, vol.~38, no.~7, 2024, pp.
  7855--7863.

\bibitem{lu2010gait}
J.~Lu and Y.-P. Tan, ``Gait-based human age estimation,'' \emph{IEEE
  Transactions on Information Forensics and Security}, vol.~5, no.~4, pp.
  761--770, 2010.

\bibitem{zhang2022gait}
S.~Zhang, Y.~Wang, and A.~Li, ``Gait energy image-based human attribute
  recognition using two-branch deep convolutional neural network,'' \emph{IEEE
  Transactions on Biometrics, Behavior, and Identity Science}, vol.~5, no.~1,
  pp. 53--63, 2023.

\bibitem{tang2019improving}
C.~Tang, L.~Sheng, Z.~Zhang, and X.~Hu, ``Improving pedestrian attribute
  recognition with weakly-supervised multi-scale attribute-specific
  localization,'' in \emph{ICCV}, 2019, pp. 4997--5006.

\bibitem{chen2019temporal}
Z.~Chen, A.~Li, and Y.~Wang, ``A temporal attentive approach for video-based
  pedestrian attribute recognition,'' in \emph{PRCV}.\hskip 1em plus 0.5em
  minus 0.4em\relax Springer, 2019, pp. 209--220.

\bibitem{wang2024spatio}
X.~Wang, Q.~Zhu, J.~Jin, J.~Zhu, F.~Wang, B.~Jiang, Y.~Wang, and Y.~Tian,
  ``Spatio-temporal side tuning pre-trained foundation models for video-based
  pedestrian attribute recognition,'' \emph{Computer Vision and Image
  Understanding}, vol. 263, p. 104588, 2026.

\bibitem{cheng2022simple}
X.~Cheng, M.~Jia, Q.~Wang, and J.~Zhang, ``A simple visual-textual baseline for
  pedestrian attribute recognition,'' \emph{IEEE Transactions on Circuits and
  Systems for Video Technology}, vol.~32, no.~10, pp. 6994--7004, 2022.

\bibitem{zhu2023learning}
J.~Zhu, J.~Jin, Z.~Yang, X.~Wu, and X.~Wang, ``Learning {CLIP} guided
  visual-text fusion transformer for video-based pedestrian attribute
  recognition,'' in \emph{CVPR Workshops}, 2023, pp. 2626--2629.

\bibitem{wu2025enhanced}
J.~Wu, Y.~Huang, M.~Gao, Y.~Niu, Y.~Chen, and Q.~Wu, ``Enhanced visual-semantic
  interaction with tailored prompts for pedestrian attribute recognition,'' in
  \emph{CVPR}, 2025, pp. 9570--9579.

\bibitem{teepe2021gaitgraph}
T.~Teepe, A.~Khan, J.~Gilg, F.~Herzog, S.~H{\"o}rmann, and G.~Rigoll,
  ``{GaitGraph}: Graph convolutional network for skeleton-based gait
  recognition,'' in \emph{ICIP}, 2021, pp. 2314--2318.

\bibitem{fu2023gpgait}
Y.~Fu, S.~Meng, S.~Hou, X.~Hu, and Y.~Huang, ``{GPGait}: Generalized pose-based
  gait recognition,'' in \emph{ICCV}, 2023, pp. 19\,595--19\,604.

\bibitem{guo2023physics}
H.~Guo and Q.~Ji, ``Physics-augmented autoencoder for {3D} skeleton-based gait
  recognition,'' in \emph{ICCV}, 2023, pp. 19\,627--19\,638.

\bibitem{lang2025beyond}
W.~Lang, S.~Hou, and Y.~Huang, ``Beyond sparse keypoints: Dense pose modeling
  for robust gait recognition,'' in \emph{ACM MM}, 2025, pp. 669--678.

\bibitem{fan2024skeletongait}
C.~Fan, J.~Ma, D.~Jin, C.~Shen, and S.~Yu, ``{SkeletonGait}: Gait recognition
  using skeleton maps,'' in \emph{AAAI}, vol.~38, no.~2, 2024, pp. 1662--1669.

\bibitem{chao2019gaitset}
H.~Chao, Y.~He, J.~Zhang, and J.~Feng, ``{GaitSet}: Regarding gait as a set for
  cross-view gait recognition,'' in \emph{AAAI}, vol.~33, no.~01, 2019, pp.
  8126--8133.

\bibitem{hou2025gaitasset}
S.~Hou, C.~Wang, A.~Li, J.~Wang, L.~Wang, and Y.~Huang, ``{GaitAsset}: In
  defense of regarding gait as a set,'' \emph{IEEE Transactions on Information
  Forensics and Security}, vol.~20, pp. 12\,301--12\,316, 2025.

\bibitem{hou2025gaitsnippet}
S.~Hou, C.~Wang, W.~Lang, Z.~Lan, and Y.~Huang, ``{GaitSnippet}: Gait
  recognition beyond unordered sets and ordered sequences,'' in \emph{ICLR},
  2026.

\bibitem{fan2020gaitpart}
C.~Fan, Y.~Peng, C.~Cao, X.~Liu, S.~Hou, J.~Chi, Y.~Huang, Q.~Li, and Z.~He,
  ``{GaitPart}: Temporal part-based model for gait recognition,'' in
  \emph{CVPR}, 2020, pp. 14\,225--14\,233.

\bibitem{lin2021gait}
B.~Lin, S.~Zhang, and X.~Yu, ``Gait recognition via effective global-local
  feature representation and local temporal aggregation,'' in \emph{ICCV},
  2021, pp. 14\,648--14\,656.

\bibitem{wang2023hierarchical}
L.~Wang, B.~Liu, F.~Liang, and B.~Wang, ``Hierarchical spatio-temporal
  representation learning for gait recognition,'' in \emph{ICCV}, 2023, pp.
  19\,582--19\,592.

\bibitem{ma2024learning}
K.~Ma, Y.~Fu, C.~Cao, S.~Hou, Y.~Huang, and D.~Zheng, ``Learning visual prompt
  for gait recognition,'' in \emph{CVPR}, 2024, pp. 593--603.

\bibitem{huang2024occluded}
P.~Huang, Y.~Peng, S.~Hou, C.~Cao, X.~Liu, Z.~He, and Y.~Huang, ``Occluded gait
  recognition with mixture of experts: An action detection perspective,'' in
  \emph{ECCV}, 2024, pp. 380--397.

\bibitem{dou2023gaitgci}
H.~Dou, P.~Zhang, W.~Su, Y.~Yu, Y.~Lin, and X.~Li, ``{GaitGCI}: Generative
  counterfactual intervention for gait recognition,'' in \emph{CVPR}, 2023, pp.
  5578--5588.

\bibitem{xiong2024causality}
H.~Xiong, B.~Feng, X.~Wang, and W.~Liu, ``Causality-inspired discriminative
  feature learning in triple domains for gait recognition,'' in \emph{ECCV},
  2024, pp. 251--270.

\bibitem{lin2019improving}
Y.~Lin, L.~Zheng, Z.~Zheng, Y.~Wu, Z.~Hu, C.~Yan, and Y.~Yang, ``Improving
  person re-identification by attribute and identity learning,'' \emph{Pattern
  Recognition}, vol.~95, pp. 151--161, 2019.

\bibitem{chen2021explainable}
X.~Chen, X.~Liu, W.~Liu, X.-P. Zhang, Y.~Zhang, and T.~Mei, ``Explainable
  person re-identification with attribute-guided metric distillation,'' in
  \emph{ICCV}, 2021, pp. 11\,813--11\,822.

\bibitem{huang2024attribute}
Y.~Huang, Z.~Zhang, Q.~Wu, Y.~Zhong, and L.~Wang, ``Attribute-guided pedestrian
  retrieval: Bridging person {Re-ID} with internal attribute variability,'' in
  \emph{CVPR}, 2024, pp. 17\,689--17\,699.

\bibitem{radford2021learning}
A.~Radford, J.~W. Kim, C.~Hallacy, A.~Ramesh, G.~Goh, S.~Agarwal, G.~Sastry,
  A.~Askell, P.~Mishkin, J.~Clark \emph{et~al.}, ``Learning transferable visual
  models from natural language supervision,'' in \emph{ICML}, 2021, pp.
  8748--8763.

\bibitem{tschannen2025siglip}
M.~Tschannen, A.~Gritsenko, X.~Wang, M.~F. Naeem, I.~Alabdulmohsin,
  N.~Parthasarathy, T.~Evans, L.~Beyer, Y.~Xia, B.~Mustafa \emph{et~al.},
  ``{SigLIP} 2: Multilingual vision-language encoders with improved semantic
  understanding, localization, and dense features,'' \emph{arXiv preprint
  arXiv:2502.14786}, 2025.

\bibitem{zhou2022learning}
K.~Zhou, J.~Yang, C.~C. Loy, and Z.~Liu, ``Learning to prompt for
  vision-language models,'' \emph{International journal of computer vision},
  vol. 130, no.~9, pp. 2337--2348, 2022.

\bibitem{pratt2023does}
S.~Pratt, I.~Covert, R.~Liu, and A.~Farhadi, ``What does a platypus look like?
  generating customized prompts for zero-shot image classification,'' in
  \emph{ICCV}, 2023, pp. 15\,645--15\,655.

\bibitem{tong2022videomae}
Z.~Tong, Y.~Song, J.~Wang, and L.~Wang, ``{VideoMAE}: Masked autoencoders are
  data-efficient learners for self-supervised video pre-training,''
  \emph{NeurIPS}, vol.~35, pp. 10\,078--10\,093, 2022.

\bibitem{wang2021actionclip}
M.~Wang, J.~Xing, and Y.~Liu, ``{ActionCLIP}: A new paradigm for video action
  recognition,'' \emph{arXiv preprint arXiv:2109.08472}, 2021.

\bibitem{zhu2023motionbert}
W.~Zhu, X.~Ma, Z.~Liu, L.~Liu, W.~Wu, and Y.~Wang, ``{MotionBERT}: A unified
  perspective on learning human motion representations,'' in \emph{ICCV}, 2023,
  pp. 15\,085--15\,099.

\bibitem{fu2021unsupervised}
D.~Fu, D.~Chen, J.~Bao, H.~Yang, L.~Yuan, L.~Zhang, H.~Li, and D.~Chen,
  ``Unsupervised pre-training for person re-identification,'' in \emph{CVPR},
  2021, pp. 14\,750--14\,759.

\bibitem{zhu2022pass}
K.~Zhu, H.~Guo, T.~Yan, Y.~Zhu, J.~Wang, and M.~Tang, ``{PASS}: Part-aware
  self-supervised pre-training for person re-identification,'' in \emph{ECCV},
  2022, pp. 198--214.

\bibitem{chen2023beyond}
W.~Chen, X.~Xu, J.~Jia, H.~Luo, Y.~Wang, F.~Wang, R.~Jin, and X.~Sun, ``Beyond
  appearance: A semantic controllable self-supervised learning framework for
  human-centric visual tasks,'' in \emph{CVPR}, 2023, pp. 15\,050--15\,061.

\bibitem{li2023clip}
S.~Li, L.~Sun, and Q.~Li, ``{CLIP-ReID}: exploiting vision-language model for
  image re-identification without concrete text labels,'' in \emph{AAAI},
  vol.~37, no.~1, 2023, pp. 1405--1413.

\bibitem{fan2023learning}
C.~Fan, S.~Hou, J.~Wang, Y.~Huang, and S.~Yu, ``Learning gait representation
  from massive unlabelled walking videos: A benchmark,'' \emph{IEEE
  Transactions on Pattern Analysis and Machine Intelligence}, vol.~45, no.~12,
  pp. 14\,920--14\,937, 2023.

\bibitem{yang2025bridging}
S.~Yang, J.~Wang, S.~Hou, X.~Liu, C.~Cao, L.~Wang, and Y.~Huang, ``Bridging
  gait recognition and large language models sequence modeling,'' in
  \emph{CVPR}, 2025, pp. 3460--3469.

\bibitem{huang2025vocabulary}
P.~Huang, S.~Hou, C.~Cao, X.~Liu, and Y.~Huang, ``Vocabulary-guided gait
  recognition,'' in \emph{NeurIPS}, 2025, pp. 155\,316--155\,338.

\bibitem{wu2026language}
Z.~Wu, C.~Zhang, S.~Jiang, H.~Xu, Z.~Liao, L.~Zhang, H.~Li, P.~Jiao, and
  H.~Wang, ``Language-guided and motion-aware gait representation for
  generalizable recognition,'' in \emph{AAAI}, vol.~40, no.~13, 2026, pp.
  10\,871--10\,878.

\bibitem{wang2022pedestrian}
X.~Wang, S.~Zheng, R.~Yang, A.~Zheng, Z.~Chen, J.~Tang, and B.~Luo,
  ``Pedestrian attribute recognition: A survey,'' \emph{Pattern Recognition},
  vol. 121, p. 108220, 2022.

\bibitem{who2000obesity}
{World Health Organization}, ``Obesity: Preventing and managing the global
  epidemic: Report of a {WHO} consultation,'' World Health Organization, {WHO}
  Technical Report Series 894, 2000.

\bibitem{hill2019effects}
A.~Hill and J.~Nantel, ``The effects of arm swing amplitude and lower-limb
  asymmetry on gait stability,'' \emph{PLOS ONE}, vol.~14, no.~12, p. e0218644,
  2019.

\bibitem{zhang2022bytetrack}
Y.~Zhang, P.~Sun, Y.~Jiang, D.~Yu, F.~Weng, Z.~Yuan, P.~Luo, W.~Liu, and
  X.~Wang, ``{ByteTrack}: Multi-object tracking by associating every detection
  box,'' in \emph{ECCV}, 2022, pp. 1--21.

\bibitem{liu2021paddleseg}
Y.~Liu, L.~Chu, G.~Chen, Z.~Wu, Z.~Chen, B.~Lai, and Y.~Hao, ``{PaddleSeg}: A
  high-efficient development toolkit for image segmentation,'' \emph{arXiv
  preprint arXiv:2101.06175}, 2021.

\bibitem{xu2022vitpose}
Y.~Xu, J.~Zhang, Q.~Zhang, and D.~Tao, ``{ViTPose}: Simple vision transformer
  baselines for human pose estimation,'' \emph{NeurIPS}, vol.~35, pp.
  38\,571--38\,584, 2022.

\bibitem{huang2019deep}
C.~Huang, Y.~Li, C.~C. Loy, and X.~Tang, ``Deep imbalanced learning for face
  recognition and attribute prediction,'' \emph{IEEE Transactions on Pattern
  Analysis and Machine Intelligence}, vol.~42, no.~11, pp. 2781--2794, 2020.

\bibitem{oquab2023dinov2}
M.~Oquab, T.~Darcet, T.~Moutakanni, H.~Vo, M.~Szafraniec, V.~Khalidov,
  P.~Fernandez, D.~Haziza, F.~Massa, A.~El-Nouby \emph{et~al.}, ``{DINOv2}:
  Learning robust visual features without supervision,'' \emph{Transactions on
  Machine Learning Research}, 2024.

\bibitem{kirillov2023segment}
A.~Kirillov, E.~Mintun, N.~Ravi, H.~Mao, C.~Rolland, L.~Gustafson, T.~Xiao,
  S.~Whitehead, A.~C. Berg, W.-Y. Lo, P.~Doll{\'a}r, and R.~Girshick, ``Segment
  anything,'' in \emph{ICCV}, 2023, pp. 4015--4026.

\bibitem{qwen2026qwen35}
{Qwen Team}, ``{Qwen3.5}: Towards native multimodal agents,''
  \url{https://qwen.ai/blog?id=qwen3.5}, 2026.

\bibitem{zhai2023sigmoid}
X.~Zhai, B.~Mustafa, A.~Kolesnikov, and L.~Beyer, ``Sigmoid loss for language
  image pre-training,'' in \emph{ICCV}, 2023, pp. 11\,975--11\,986.

\bibitem{wu2020person}
J.~Wu, H.~Liu, J.~Jiang, M.~Qi, B.~Ren, X.~Li, and Y.~Wang, ``Person attribute
  recognition by sequence contextual relation learning,'' \emph{IEEE
  Transactions on Circuits and Systems for Video Technology}, vol.~30, no.~10,
  pp. 3398--3412, 2020.

\bibitem{teepe2022towards}
T.~Teepe, J.~Gilg, F.~Herzog, S.~H{\"o}rmann, and G.~Rigoll, ``Towards a deeper
  understanding of skeleton-based gait recognition,'' in \emph{CVPR Workshops},
  2022, pp. 1569--1577.

\bibitem{zhang2023spatial}
C.~Zhang, X.-P. Chen, G.-Q. Han, and X.-J. Liu, ``Spatial transformer network
  on skeleton-based gait recognition,'' \emph{Expert Systems}, vol.~40, no.~6,
  p. e13244, 2023.

\bibitem{meng2025fastposegait}
S.~Meng, Y.~Fu, S.~Hou, X.~Hu, C.~Cao, X.~Liu, and Y.~Huang, ``From
  fastposegait to gpgait++: Bridging the past and future for pose-based gait
  recognition,'' \emph{IEEE Transactions on Pattern Analysis and Machine
  Intelligence}, vol.~47, no.~9, pp. 8231--8248, 2025.

\bibitem{fan2023exploring}
C.~Fan, S.~Hou, Y.~Huang, and S.~Yu, ``Exploring deep models for practical gait
  recognition,'' \emph{arXiv preprint arXiv:2303.03301}, 2023.

\bibitem{zhou2023solution}
Y.~Zhou, H.-M. Hu, J.~Yu, Z.~Xu, W.~Lu, and Y.~Cao, ``A solution to
  co-occurence bias: attributes disentanglement via mutual information
  minimization for pedestrian attribute recognition,'' in \emph{IJCAI}, 2023,
  pp. 1831--1839.

\end{thebibliography}
\end{document}